\documentclass{article}

\usepackage{microtype}
\usepackage{graphicx}
\usepackage{subcaption}
\usepackage{booktabs} % for professional tables
\usepackage{amsmath}
\usepackage{epstopdf}
\usepackage{bm,amsmath,amssymb,amsthm}
\usepackage{Definitions}
\usepackage{enumitem}
\usepackage{bookmark}

\newtheorem{problem}{Problem}
\RequirePackage{fancyhdr}
\RequirePackage{color}
\RequirePackage{algorithm}
\RequirePackage{algorithmic}
\RequirePackage{natbib}
\RequirePackage{eso-pic} % used by \AddToShipoutPicture 
\RequirePackage{forloop}
\usepackage[letterpaper, portrait, margin=1in]{geometry}
\usepackage{authblk}
\usepackage{enumitem}
\usepackage[symbol]{footmisc}
\title{Beyond Convolutions: A Novel Deep Learning Approach for\\ Raw Seismic Data Ingestion}

\author[1]{Zhaozhuo Xu*}
\author[1]{Aditya Desai*}
\author[2]{Menal Gupta}
\author[2]{Anu Chandran}
\author[2]{Antoine Vial-Aussavy}
\author[1]{Anshumali Shrivastava}
\affil[1]{Department of Computer Science,  Rice University}
\affil[2]{Shell International E$\&$P, Inc}

\affil[ ]{\textit {zx22,apd10,anshumali@rice.edu} \qquad \textit { Menal.Gupta,Anu.Chandran,a.vial-aussavy@shell.com}}

\date{}

\usepackage{titlesec}

\titleformat*{\section}{\large\bfseries\raggedright}
\titleformat*{\subsection}{\normalsize\bfseries\raggedright}
\titleformat*{\subsubsection}{\normalsize\scshape\raggedright}

\setlist{leftmargin=1.8em}

\begin{document}
\setlength{\columnsep}{16pt}
\twocolumn[
\maketitle

% \vskip 0.3in
]

\renewcommand{\thefootnote}{\fnsymbol{footnote}}
\footnotetext[1]{Equal contributions.}

\begin{abstract}
Traditional seismic processing workflows (SPW)  are expensive, requiring over a year of human and computational effort.  Deep learning (DL) based data-driven seismic workflows (DSPW) hold the potential to reduce these timelines to a few minutes. Raw seismic data (terabytes) and required subsurface prediction (gigabytes) are enormous. This large-scale,  spatially irregular time-series data poses seismic data ingestion (SDI) as an unconventional yet fundamental problem in DSPW.  Current DL research is limited to small-scale simplified synthetic datasets as they treat seismic data like images and process them with convolution networks. Real seismic data,  however,  is at least 5D. Applying 5D convolutions to this scale is computationally prohibitive. Moreover, raw seismic data is highly unstructured and hence inherently non-image like.  We propose a fundamental shift to move away from convolutions and introduce SESDI: Set Embedding based SDI approach.  SESDI first breaks down the mammoth task of large-scale prediction into an efficient compact auxiliary task. SESDI gracefully incorporates irregularities in data with its novel model architecture. We believe SESDI is the first successful demonstration of end-to-end learning on real seismic data. SESDI achieves SSIM of over 0.8 on velocity inversion task on real proprietary data from the Gulf of Mexico and outperforms the state-of-the-art U-Net model on synthetic datasets.

% \vspace{-0.5cm}
\end{abstract}
\vspace{-3mm}
\section{Introduction}
\begin{figure*}
\begin{subfigure}{.65\textwidth}
  \centering
  \includegraphics[width=.9\linewidth]{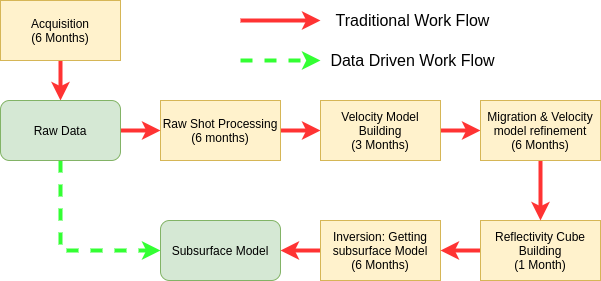}
%   \caption{Traditional workflow involved in seismic processing of raw data to getting a subsurface model. As can be noticed traditional workflow requires approximately over an year time to generate a subsurface image. Data-driven workflows try to bypass this tedious process by directly predicting subsurface model from raw data}
\caption{}
  \label{fig:workflow}
\end{subfigure}
\begin{subfigure}{.3\textwidth}
  \centering
  \includegraphics[width=1\linewidth]{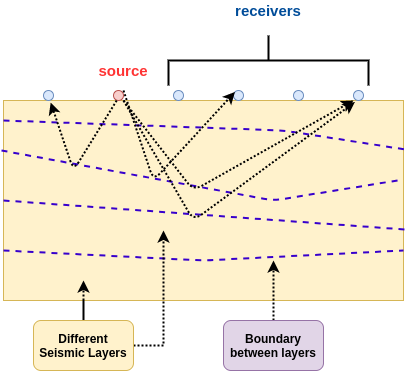}
%   \caption{Reflection Seismology for collecting seismic data: Mechanical Perturbation (sound wave) created at source, propagates through layers of earth and reflects subsurface boundaries. The receivers record amplitude time-series information.}
\caption{}
  \label{fig:refseis}
\end{subfigure}
\vspace{-0.2cm}
\caption{\textbf{Left}:Traditional workflow involved in seismic processing of raw data to getting a subsurface model. It requires approximately over an year time to generate a subsurface image. Data-driven workflows try to bypass this tedious process by directly predicting subsurface model from raw data. \textbf{Right}: Reflection Seismology for collecting seismic data: Mechanical Perturbation (sound wave) created at source, propagates through layers of earth and reflects subsurface boundaries. The receivers record amplitude time-series information.}
\label{fig:workflow_refseis}
\vspace{-0.5cm}
\end{figure*}

    % need some introduction to the problem. FWI/VMB may not be known in community. 
     Seismic reflection data, obtained via reflection seismology ~\cite{waters1981reflection}, is a dominant information source to understand the earth's subsurface. Figure \ref{fig:refseis} shows a typical setup for reflection seismology. It uses apparatus such as air guns (sources) to generate pressure waves on the earth's surface. These waves encounter geological contrast in the subsurface and reflect back to the multiple geophones (receivers) positioned on the surface. This time-dependent recording of amplitudes by receivers constitutes the seismic data. The recording of each source, receiver pair is called a trace. The positions(x,y coordinates) of sources and receivers are referred to as acquisition geometry. Each trace is associated with additional metadata such as the signature of disturbance, etc. Even ignoring this metadata, the trace is at least five-dimensional (surface coordinates for both source and receiver and time). A typical seismic survey is extensive (100 km $\times$ 100km), with each 1km x 1km patch having tens of thousands of traces resulting in data sizes of 10-100 terabytes. Conventionally, seismic data is processed using wave propagation physics to invert for relevant earth properties. We present a novel Deep learning based SDI approach, denoted as Set Embedding based SDI (SESDI), that efficiently ingests raw seismic data directly with applicability in any seismic processing workflow. This paper selects seismic velocity inversion, one of the most crucial subsurface properties,  as a case study for SESDI. SESDI benefits effective analysis of seismic data in environmental applications such as earthquake understanding~\cite{sergeant2016complex,julian1996earthquake,arora2010global}, groundwater investigation~\cite{zelt20063d} and hydrocarbon exploration~\cite{alsadi2017seismic}.

% \anshu{Raw data to migration --  which is mostly forward modeling  ---  (but we need extra image) --  Image of reflectors  -- NOT AN EASY PROCESS -- a few week to a few months. UNet model only comes in play after 6 months.}
 \textbf{Expensive seismic processing workflows:} Traditional state-of-the-art methods of processing seismic data consume significant resources and time~\cite{onajite2013seismic}.  Figure \ref{fig:workflow} shows broader steps in traditional workflow commonly used in the ``Big Oil" industry.  It is well known~\cite{le2012advanced} that each of the sequential steps in traditional workflows require several months. This time inefficiency is primarily due to their dependence on human expertise to process vast amounts of raw data.  With the increasingly comprehensive and complex seismic data generated by advanced sensing technology, traditional workflows prove to be more expensive than ever. Hence, there is an urgent need to devise techniques to reduce human involvement in seismic processing. 

     \textbf{Promise of deep Learning (DL):} The data-driven approaches ~\cite{fichtner2010full,lin2014acoustic,lin2014ultrasound,lin2015quantifying} are progressively sought alternative to improve seismic workflows' efficiency. Data-driven deep learning (DL) approaches aim to learn a non-linear mapping from the raw seismic data to the final subsurface image evading most of the time-consuming sequential steps in traditional workflows.  With the fast inference of DL models, data-driven approaches can potentially reduce workflows' duration from months to minutes. Therefore, it is not surprising that DL based workflows in seismic processing are getting the attention and investment they deserve.
    ~\cite{dramsch2018deep,li2019generative,griffith2019deep, zheng2019applications,Yang2019Deeplearning,sun2020deep,inversionnet,rojas2020physics,faultmauricio,semblance}
    \textbf{Convolutions in data-driven processing} Treating parts of raw seismic data as images is standard in the oil industry and geophysics community. Undoubtedly, image is the most convenient and interpretable representation of complex data. The recent data-driven research , inheriting the image-bias from traditional practices and inspired by the success of convolutions in computer vision, attempts to process the raw seismic data as images and apply convolution based networks to ingest them. ~\cite{inversionnet,dramsch2018deep,di2018using,zheng2019applications,Yang2019Deeplearning,sun2020deep,rojas2020physics}.

\begin{figure*}
\begin{subfigure}{.6\textwidth}
  \centering
  \includegraphics[width=.98\linewidth]{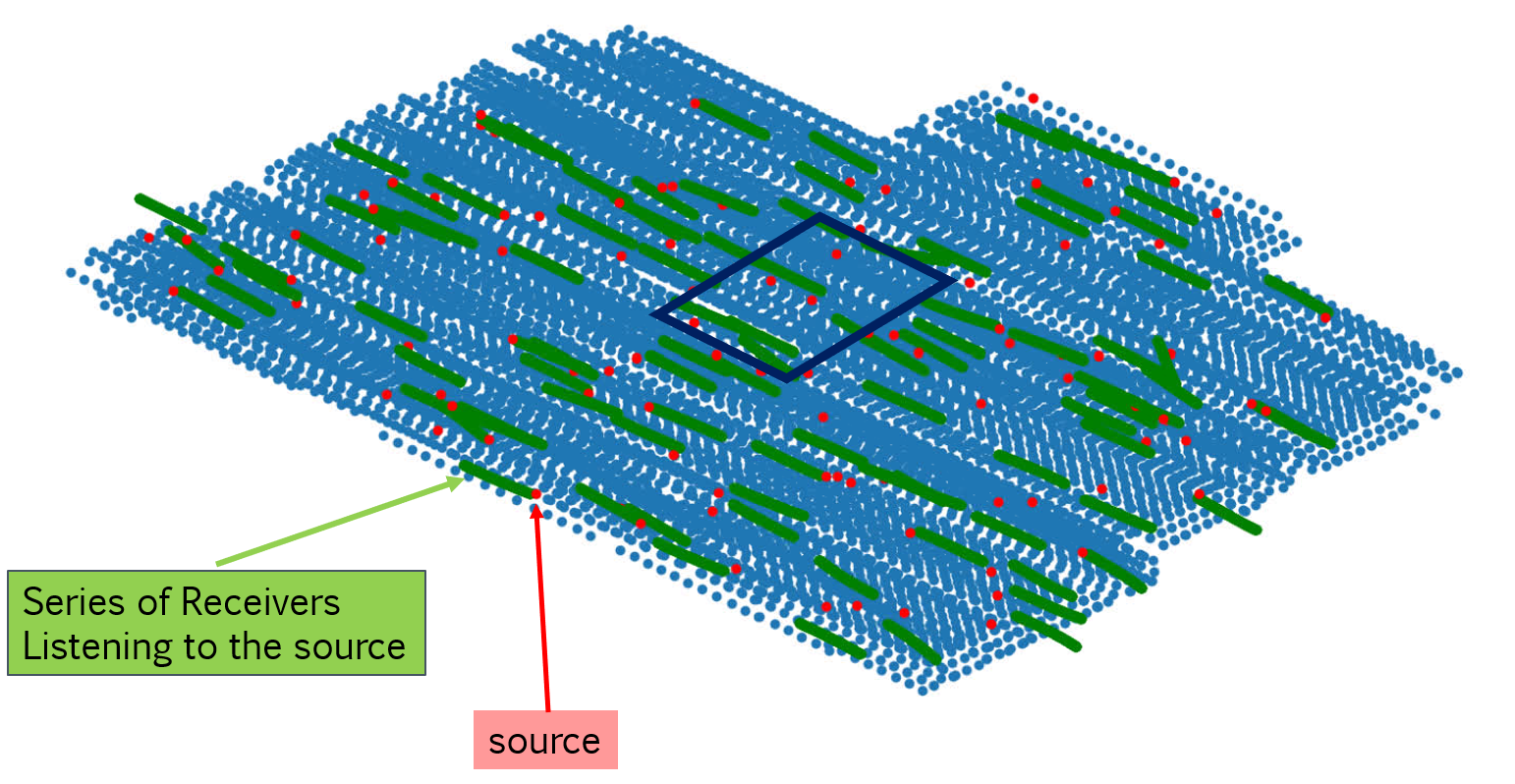}
%   \caption{A top view of a typical real survey. The survey area is large (around 100Km $\times$ 100 Km) and the streamer (in marine survey) moves across the area collecting data at various places. The blue dots represent all the source points in the survey. Red dots and nearby green segments refer to a pair of source and series of receivers. Typically we want to focus of a particular area to predict the subsurface properties. }
  \label{fig:sfig1}
\end{subfigure}
\begin{subfigure}{.38\textwidth}
  \centering
  \includegraphics[width=1.06\linewidth]{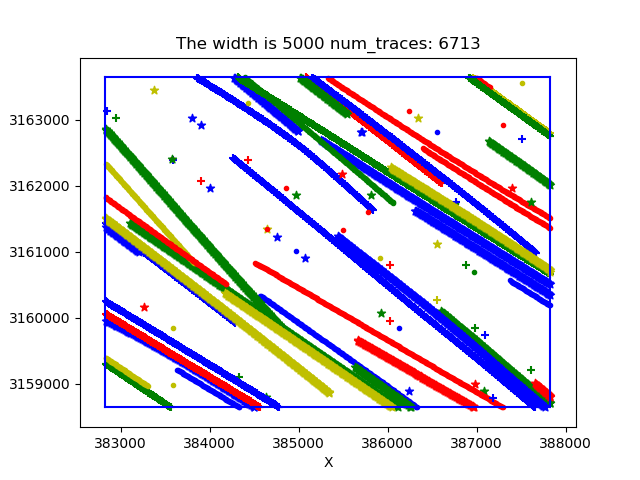}
%   \caption{Closer look at the acquisition over 5Km x 5Km area. Different colors and symbols are used to identify pairs of source and receiver lines. Figure shows that real acquisition is highly irregular emphasizing the oversimplification of data used in current models.}
  \label{fig:sfig2}
\end{subfigure}
% \caption{While performing a seismic survey, a set of sources and receivers move over the area of interest and shots are fired and recorded at different locations. This figure is drawn from a real survey and shows how irregular the acquisition is}
\caption{Irregular Acquisition in real SDI. \textbf{Left}: A top view of a real survey. The survey area is large (around 100Km $\times$ 100 Km) and the streamer (in marine survey) moves across the area to collect data. The blue dots represent the source points in the survey. Red dots and nearby green segments refer to a pair of source and series of receivers. We focus on a particular area to predict the subsurface properties. \textbf{Right}: Details at the acquisition over a 5Km x 5Km area. Different colors and symbols are used to identify pairs of source and receiver lines. The real acquisition is highly irregular emphasizing the oversimplification of data used in current convolution models. Seismic data courtesy of TGS.}
\label{fig:acquisition}
\end{figure*}

\textbf{Limited success of deep learning in real seismic processing:}
Current DL based SDI with CNN architecture suffers severely from two major constraints 1) Real seismic data is at least 5 dimensional. Applying 5D convolutions to the extensive  seismic data is computationally infeasible. 2) Real seismic data collection is highly irregular and unstructured (see Figure \ref{fig:acquisition}). However, CNNs' strong requirements on structured data limit their application in regular consistent acquisition geometry. Thus, to the best of our knowledge, most CNN based SDI approaches are prohibitive to real seismic settings. These methods' current success is produced on simulations over 2D velocity patches with regular acquisition geometry (see Figure \ref{fig:syntheticAQ}). Here, sources and receivers are uniformly distributed on the surface and each receiver listening to every source. The gap between simulation and practice is too wide to extend these methodologies to practical environmental applications.

\textbf{Major contributions in the paper}
\begin{itemize}[leftmargin=*,nosep]
    \item We propose a fundamental shift to move away from convolutions in data-driven seismic processing.
    \item We introduce SESDI approach which 1.) redefines the seismic inversion problem on terabytes of data into the auxiliary task of predicting small blocks of property models from their contextual traces. Thus making training and inference of deep learning models on real data feasible. 2) We provide a deep semantic embedding model for a set of contextual traces that is robust to highly irregular acquisition geometry. 3) SESDI approach is a SDI approach; the embedding obtained for a context can be used for any downstream task such as velocity inversion. 
    \item We believe, SESDI is the first-ever data-driven workflow that can reasonably ingest real data and provide geologically sound results. In the experiments on real proprietary data from the survey on the Gulf of Mexico, we achieve a SSIM similarity (0-worst, 1-best) of over 0.8 on predicting 3D velocity model cubes.
    \item We outperform state-of-the-art U-Net model with just MLP functional blocks on synthetic datasets,  suggesting re-thinking the need for convolutions for SDI.
\end{itemize}

\section{Background}
% \vspace{-0.2cm}

\subsection{Seismic Data and Associated Applications} % @zhaozhuo Seismic data ingestion means a "way to consume/process data"
% \vspace{-0.2cm}

For environmental applications such as hydrocarbon exploration and groundwater investigation, geophysicists aim to create a subsurface property model. Generally, the property model map is stored as a discretized 3d cube with the uniform value per partition.

\begin{definition}[\textbf{Subsurface Property Model}]
Subsurface property model is defined as a map $\mathcal{P}:R^3 \rightarrow R^k$ which maps each point in the subsurface to a tuple of properties like density, shear modulus bulk modulus, etc

\end{definition}
Reflection Seismology \cite{waters1981reflection} is used to collect information about the subsurface which is used to construct property model. We restrict our discussion to the marine survey in this paper.
%In order to construct a property model, information about the subsurface is collected using a technique called as Refelction Seismology \cite{waters1981reflection}
% \vspace{-0.2cm}
\subsubsection{Reflection Seismology:} 
% \vspace{-0.2cm}

As shown in Figure \ref{fig:refseis}, a source and a set of receivers are positioned on the surface of earth. A mechanical perturbation is created using air guns at the source. The pressure waves thus generated travels through the subsurface. They refract and reflect at different layer boundaries in the subsurface. The reflected pressure waves that reach the receivers are recorded to generate a time-series data. Hence, each source shot and receiver pair is associated with a 1D time series data called the \textit{trace} which is formally defined as

\begin{definition}[\textbf{Trace Data (u)}]\label{trace}
Given a time duration t, and a frequency f, a trace at a location x on surface is defined as the sequence $\{ u(t_i, x) \}_{i=1}^{tf}$ of acoustic pressure sensed at x at time $t_i$,  where $t_i$ is the $i^{th}$ time sample when sampling with frequency f for time t.
\end{definition}
\begin{definition}[\textbf{Geometry-Aware Trace (t)}]\label{gatrace}
The trace data, u, along with its (source,reciever) locations, q, together constitute a geometry-aware trace: $t = (u,q)$. We use functions $u=\mathcal{D}(t)$ and $q=\mathcal{A}(t)$ to extract trace data and location information respectively from the trace
\end{definition}

In a typical survey, a source is located on the streamer and long cables with equidistant receivers are attached to the streamer. The streamer moves over the area of interest and collects a large set of traces for different (source, receiver) pairs. If the total number of traces obtained from the survey is N, we define raw-data as the set of all geometry-aware traces. Specifically $\textrm{raw-data} = \{ t_i\}_{i=1}^N$. The set of (source, receiver) locations associated with the collected traces is referred to as the acquisition geometry that formally defined as

\begin{definition}[\textbf{Acquisition Geometry($\mathbb{A}$)}]\label{def:geometry}
Acquisition Geometry is defined as the set $\{\mathcal{A}(t) |  t \in \textrm{raw-data} \}$
\end{definition}
Traces are consequence of the propagation and reflection of pressure wave through the subsurface. The propagation of reflected waves is generally modelled by the Acoustic Wave Equation, defined as below
\begin{definition}[\textbf{Acoustic Wave Equation}]\label{def:waveq}
Given a position vector $x$, spatially varying wave velocity denoted as $v(x)$. Provided with a seismic source $q_s(t,x)$ located as $x_s$. The acoustic pressure $u(t,x)$ in time $t$ is determined by:
\begin{align}
    m\frac{\partial^2 u(t,x)}{\partial t^2}-\nabla^2 u(t,x)+\eta \frac{\partial u(t,x)}{\partial t}
    =q_s(t,x)
\end{align}

where $m(x)=v^{-2}(x)$ is the squared slowness at $x$, $\eta(x)$ is the space-dependent dampening parameter for the absorbing boundary layer~\cite{cerjan1985nonreflecting} and $\nabla^2$ is the laplacian operator.
\end{definition}

% \vspace{-0.2cm}
\subsubsection{Applications of Seismic Data}
% \vspace{-0.2cm}

The final aim of seismic processing is to obtain a subsurface property model. This problem is referred to as the seismic inversion problem stated formally below. This property model can then identify the rock and fluid types in the subsurface.
\begin{problem}[\bf Seismic Inversion
]\label{prob:si}
Given a region of interest $S \subset R^d$, set of traces T and associated Acquisition Geometry $\mathbb{A}$,
Seismic Inversion(SI) is defined as the problem of getting the property model $\mathcal{P}$
\vspace{-0.2cm}
\end{problem}

The property here can be any set of elastic properties, intermediate velocity or reflectivity index, etc. We use the term \textit{property} in the most general manner. The subsurface region of interest, $S$, in problem definition \ref{prob:si}can be 2D or 3D space. Accordingly, the inversion problems are categorized as 2D or 3D problems. 2D problems generally appear in synthetically generated datasets.
\subsection{Relevant Prior Work}
% \vspace{-0.2cm}

\textbf{Traditional approaches} \cite{seismicbook}: Traditionally SDI for seismic workflows comprises a series of sequential steps involving computer-assisted human analysis. Generally, these steps are physics-driven. An example of workflow with approximate timelines is shown in figure \ref{fig:workflow}. Raw shot processing cleans up the data by removing unintended noises that appear in the data. It includes sequential steps like de-noising, de-signature, de-multiple etc \cite{seismicbook, denoising1, designature1, demultiple1}. The clean data is used to generate a velocity model of the subsurface using techniques of tomography~\cite{nolet1987seismic,shapiro2005high} or full-waveform inversion~\cite{virieux2009overview,brossier2009seismic}. The velocity model is then refined iteratively by migrating the waves and computing errors like residual move out \cite{xie2008finite}. Once a refined velocity model is obtained, it is used to build a reflectivity cube of the subsurface. This cube is then inverted to obtain elastic properties of the subsurface (subsurface property model). Types of rocks and fluids in the subsurface are identified by analyzing the property Model. Each of these steps require involvement of domain experts to guide the inversion to an effective and applicable solution. These procedures extend the overall process to as lengthy as a year. For a detailed discussion on these steps, we refer the readers to \cite{seismicbook}.

\begin{figure}
    \centering
    \includegraphics[scale=0.39]{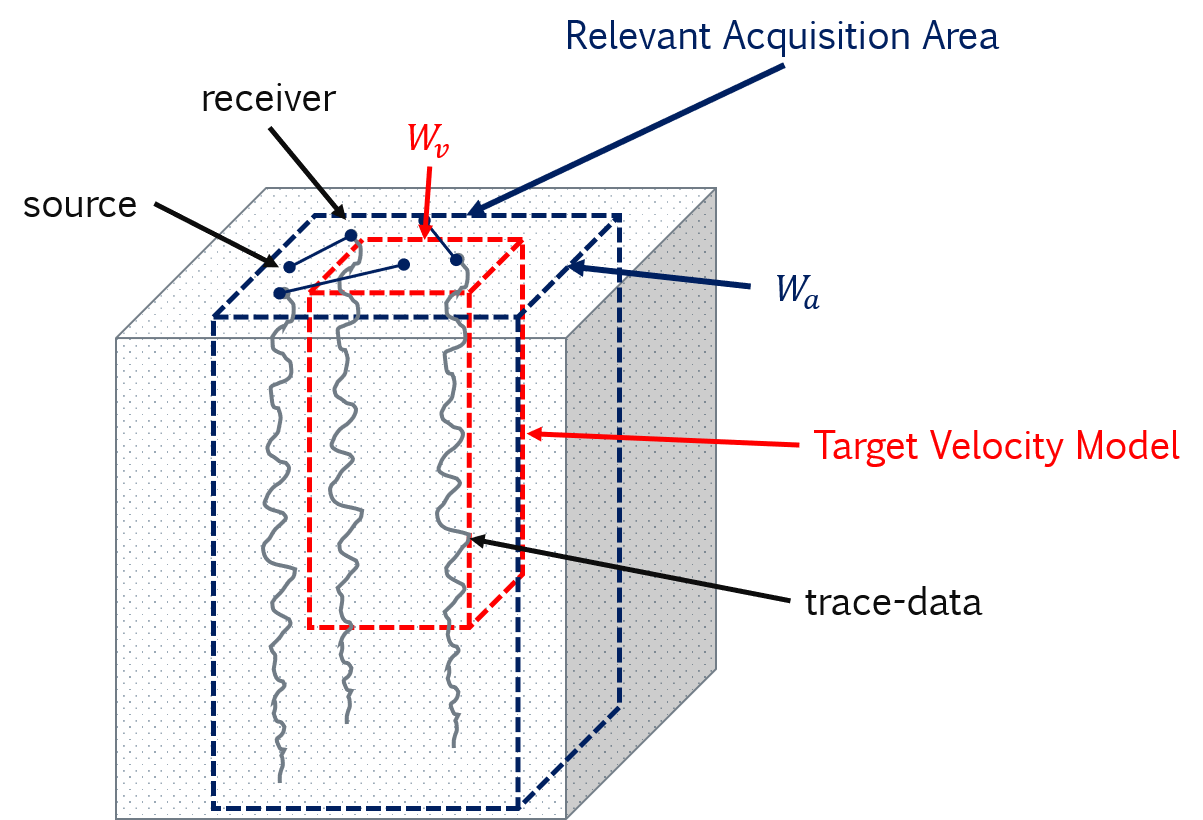}
    \caption{ Construction of context for a particular property model block placed near surface of earth.}
    \label{fig:exgen}
    \vspace{-0.6cm}
\end{figure}

\textbf{Recent DL approaches: } Though the application of data-driven methods to seismic processing dates back to \cite{nath1999velocity}, there has been renewed interest since the success of deep learning \cite{goodfellow2016deep}. The data-driven methods~\cite{inversionnet,Yang2019Deeplearning} aim to utilize the strength of deep learning ~\cite{goodfellow2016deep} to exploit the available extensive complex seismic data fully. Most existing methods formulate Problem~\ref{prob:si} as an image segmentation problem and utilize convolutional neural networks (CNN)~\cite{ronneberger2015u,chen2017deeplab}. In this formulation, the traces are combined as multi-channel images and the corresponding subsurface property model is regarded as a segmentation mask. Then CNN architectures such as U-Nets\cite{zheng2019applications,Yang2019Deeplearning,sun2020deep} or encoder-decoders~\cite{inversionnet,rojas2020physics} are applied to learn a non-linear mapping between image-mask pairs. As stated earlier, these approaches do not scale to real data.

\begin{figure}
    \centering
    \includegraphics[scale=0.26]{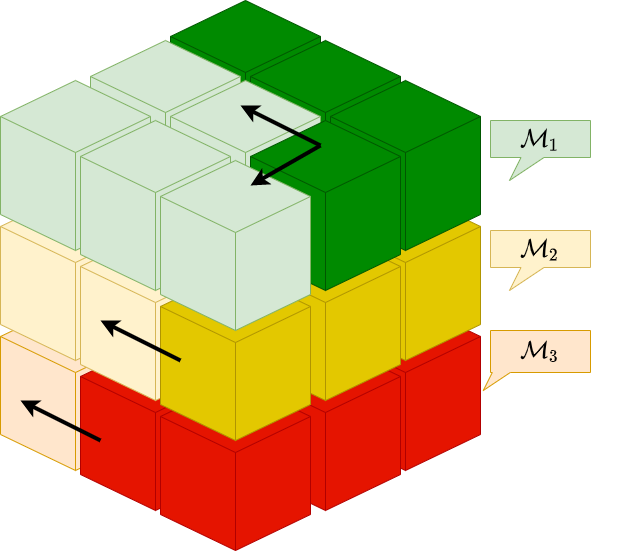}
    \caption{Creating entire property model using different learning models at different depths with each model traversing entire horizontal surface predicting one block at a time.}
    \label{fig:stiching}
    \vspace{-0.6cm}
\end{figure}

% \vspace{-0.2cm}
\section{SESDI: Set Embedding based Seismic Data Ingestion} \label{sec:approach}
% \vspace{-0.2cm}
\subsection{Decomposition of Original Problem}
% \vspace{-0.2cm}

The raw data for a typical survey measures up to several terabytes, while the discretized output property model of the subsurface beneath the survey area is in several gigabytes. It is prohibitive to learn a mapping that directly attempts to ingest terabytes of seismic data and produce a big property model. Unlike current state-of-the-art methods, SESDI is designed to work for real data. SESDI surmounts the problem of scale by breaking this task into an achievable auxiliary task of predicting small property model blocks using the contextual traces and then stitching them together to obtain the complete property model. One can draw an analogy to NLP models on machine document translation. Instead of learning a complete mapping between two documents, state-of-art techniques design a model to predict the next word given the current context and use this to translate the entire document. Similarly, SESDI predicts a property model block by considering the traces that appear around it. 

Let the property model block of size $w_p \times w_p \times d$ with its midpoint located at a particular location, say $q$, in 3D space be denoted by $\mathcal{P}(q,w_p,d)$. We define the context for such a block, shown in figure \ref{fig:exgen}, formally, 
\begin{definition}[\textbf{Context $\mathcal{C}(q, w)$}]
The context, $\mathcal{C}(q, w)$, at a point q, parameterised by width w is defined as the set of geometry-aware traces whose acquisition(source and reciever locations) lie inside $Area(q, w)$, i.e. the area $w \times w$ centered at $(q.x, q.y)$
\begin{align*}
\mathcal{C}(q, w) = \{t | &t \in \textrm{raw-data}, \mathcal{A}(t) \in Area(q, w) \}
\end{align*}
\end{definition}

SESDI aims to lean a mapping, $\mathcal{M}$, between $context(q, w_a)$ and property model block $p(q, w_p, d)$. The scale of this learning problem is much lesser as compared to the original problem, making it feasible to train a model efficiently. SESDI proposes to learn different mappings for different depths. For example if the total depth of complete property model is D and we choose to predict model blocks of depth d, then SESDI proposes to learn $\lceil D/d \rceil$ mappings $\{ \mathcal{M}_i \}_{i=1}^{\lceil D/d \rceil}$. With the help of $\lceil D/d \rceil$ models, SESDI can predict the entire property model as shown in figure \ref{fig:stiching}. Specifically, in order to predict property at a arbitrary location, say q, in the subsurface, we would use the the mapping $\mathcal{M}_{\lceil q / d \rceil}$ with context $context(q, w(\lceil q / d \rceil))$ and then locating the point q inside the predicted block. The width of context $w$ is should increase when predicting property at higher depth. Hence, we use $w(\lceil q / d \rceil)$ to make the dependence explicit.

\subsection{Learning a Mapping $\mathcal{M}$}
% \vspace{-0.2cm}

Let us focus on predicting the property model block at a specific depth.
%This section introduces the distributed vector representation of seismic traces.  @zhaozhuo i dont know what this means any reference? if this is standard term. 
% \vspace{-0.2cm}

\subsubsection{Information-aligned trace embeddings}\label{sec:tr_emb}
% \vspace{-0.1cm}

For a particular property model block, $P(q, w_p, d)$, let the context block be $\mathcal{C}(q,w)$. Let $\mathcal{C}(q,w) = \{t_i \}_{i=1}^n$. Trace data, $\mathcal{D}(t)$, is a sequential amplitude recording  of pressure wave measured at receiver. The receiver catches the reflected waves from different depths and directions. The recorded amplitude is a superposition of all waves that reach the receiver simultaneously with lesser contributions from waves reflected from faraway points. Hence, each trace captures relevant information about the entire subsurface. In this respect, the raw data is a collection of highly redundant and correlated traces containing overlapping pieces of information about the entire subsurface. SESDI has to align the information of these pieces before combining them to generate property model block.
\begin{figure}
    \centering
    \includegraphics[scale=0.6]{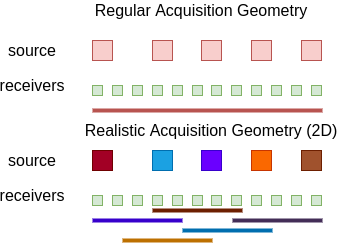}
    \caption{Regular AG used in synthetic datasets is to have sources and receivers uniformly distributed on surface and have every receiver listen to every source. Realistic AG for 2D is designed to mimic Real acquisition. Here, randomly sampled contiguous segments of receivers are made to listen to each source. The color shows the pairings of receivers and sources.}
    \label{fig:syntheticAQ}
    % \vspace{-0.5cm}
\end{figure}
The information of acquisition of a trace t, $\mathcal{A}(t)$ is important to align the semantic information from different traces correctly and consistently. In SESDI, we acheive this aligned semantic embedding by first projecting the trace-data $\mathcal{D}(t)$ and its acquisition $\mathcal{A}(t)$ independently into intermediate high dimensional space, combining them and projecting them in the final target embedding space. Formally, 
\begin{align*}
    \mathcal{E} (t) 
      & = \phi(t) =  \mathcal{F}_t \left(\left[\mathcal{F}_{aq}(\mathcal{A}(t)) , \mathcal{F}_{d}(\mathcal{D}(t))\right]\right)
\end{align*}
The resulting bag $\{\mathcal{E}(t_i)\}_{i=1}^n$ is a set of information-aligned trace embeddings. 
In our experiments on seismic data, we use Multi layer Perceptrons (MLPs) for the functional blocks $\mathcal{F}_{aq}$, $\mathcal{F}_d$ and $\mathcal{F}_t$.

% \vspace{-0.2cm}
\subsubsection{Representation of Context $\mathcal{C}$ }
% \vspace{-0.1cm}

In a real survey (Figure \ref{fig:acquisition}), the acquisition geometry  and the number of traces in a context $\mathcal{C}(q,w)$ change with changing location $q$ of the context. Due to this irregularity of traces across different contexts, it is impossible to impose specific structural formulations such as convolution or concatenation on the trace embeddings.  
To tackle these issues, we propose a set formulation to learn a high dimensional representation of the context $\mathcal{C}(q,w)$ . If $\mathcal{C}(q, w) = \{t_i\}_{i=1}^n$ then, we define the context embedding as,
\begin{align*}
\mathcal{E}({\mathcal{C}(q, w)})  = \frac{1}{n} \Sigma_{i=1}^n \phi (t_i)
\end{align*}
where $\phi(t_i)=\mathcal{E}(t_i)$ is the information-aligned embedding of the trace $t_i$ as defined in \ref{sec:tr_emb}. Alignment of trace embeddings is of key significance in this formulation.

\begin{figure}
    \centering
    \includegraphics[scale=0.33 ]{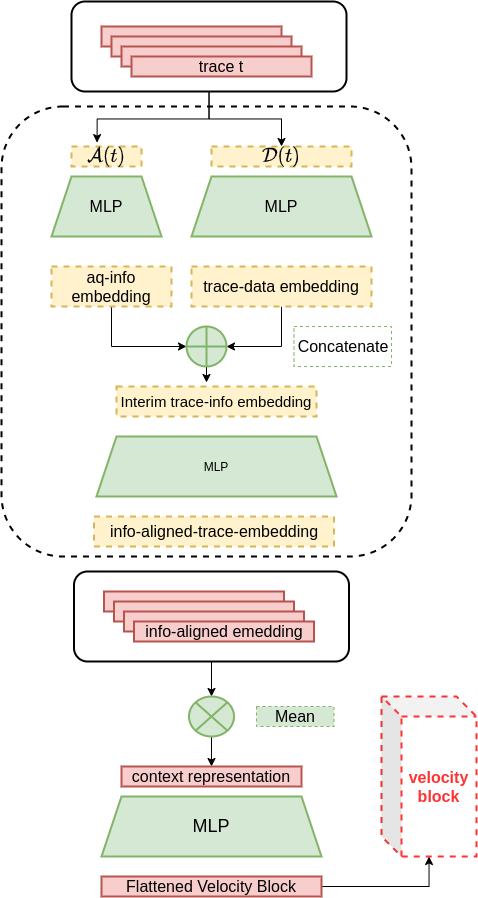}
    \caption{SESDI Model Architecture}
    \label{fig:model}
    \vspace{-0.6cm}
\end{figure}

% \vspace{-0.2cm}
\subsubsection{Downstream Tasks}
% \vspace{-0.1cm}

Once a context representation is obtained, we can now use this to predict the predict any property we desire, such velocity model, reflectivity model or even elastic properties of the subsurface. The property model is predicted by processing the context embedding using a function $\rho$.
\begin{align*}
    \hat{\mathcal{P}} = \rho(\mathcal{E}(\mathcal{C}(q,w)))
\end{align*}
 Again, we implement $\rho$ as a MLP. We learn all the functional blocks in an end-to-end fashion. Hence, 
 \begin{align*}
    \hat{\mathcal{P}} =\rho(\frac{1}{n} \Sigma_{i=1}^n \mathcal{F}_t \left(\left[\mathcal{F}_{aq}(\mathcal{A}(t_i)) , \mathcal{F}_{d}(\mathcal{D}(t_i))\right]\right)))
 \end{align*}
 Coincidentally, this formulation also agrees with a recent paper on processing sets via deep learning
\cite{deepsets}.
An overview of SESDI model is shown in Figure \ref{fig:model}

\begin{figure*}[htbp]
\centering
\begin{subfigure}{.55\textwidth}
\centering
\mbox{\hspace{-0.2in}
 \includegraphics[width=.21\textwidth]{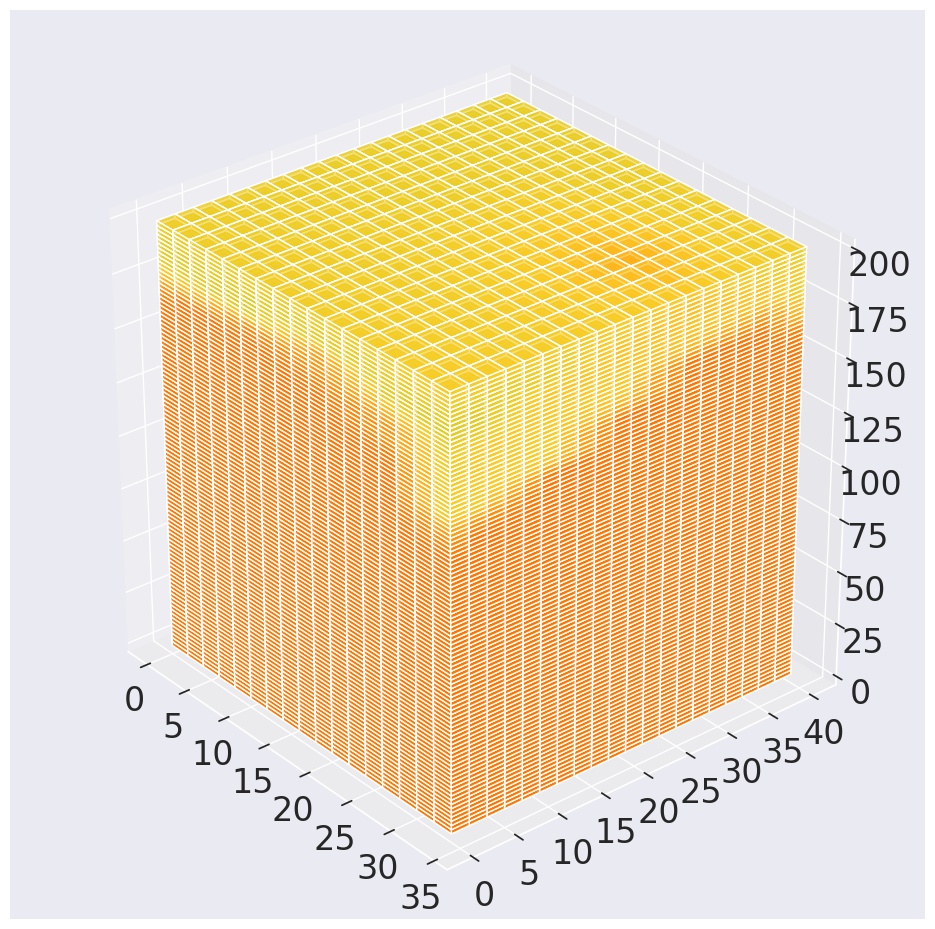}
\includegraphics[width=.21\textwidth]{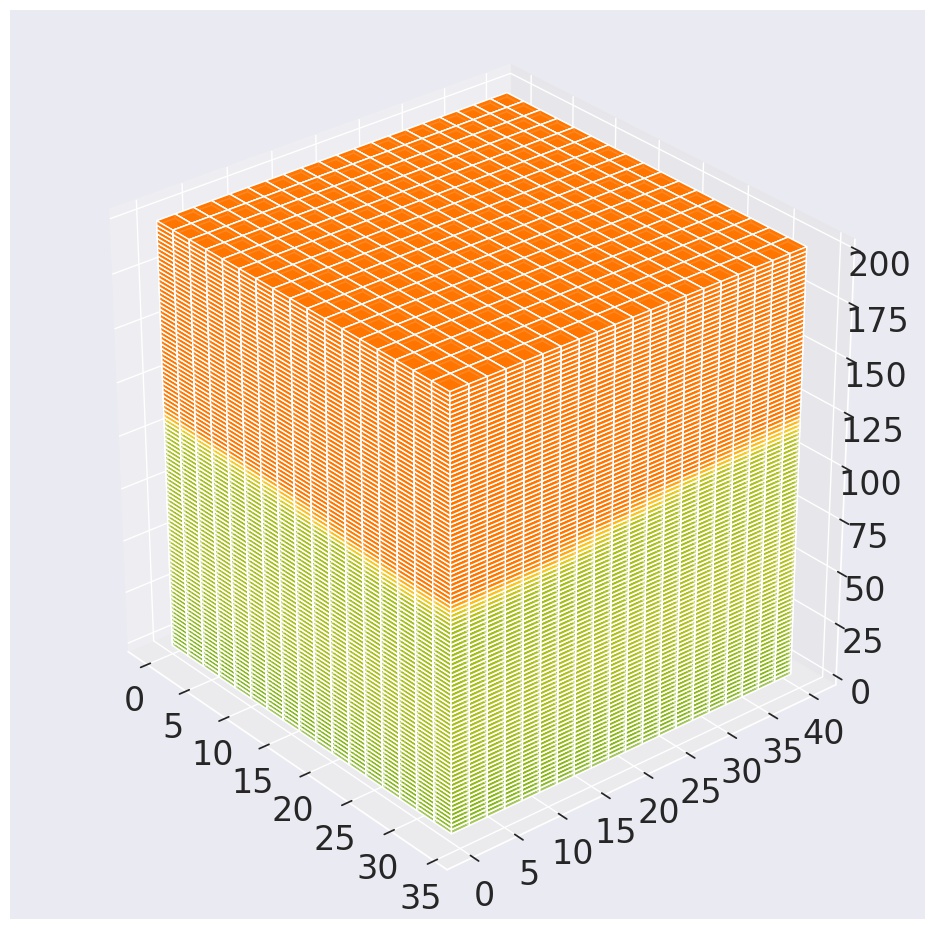}
\includegraphics[width=.21\textwidth]{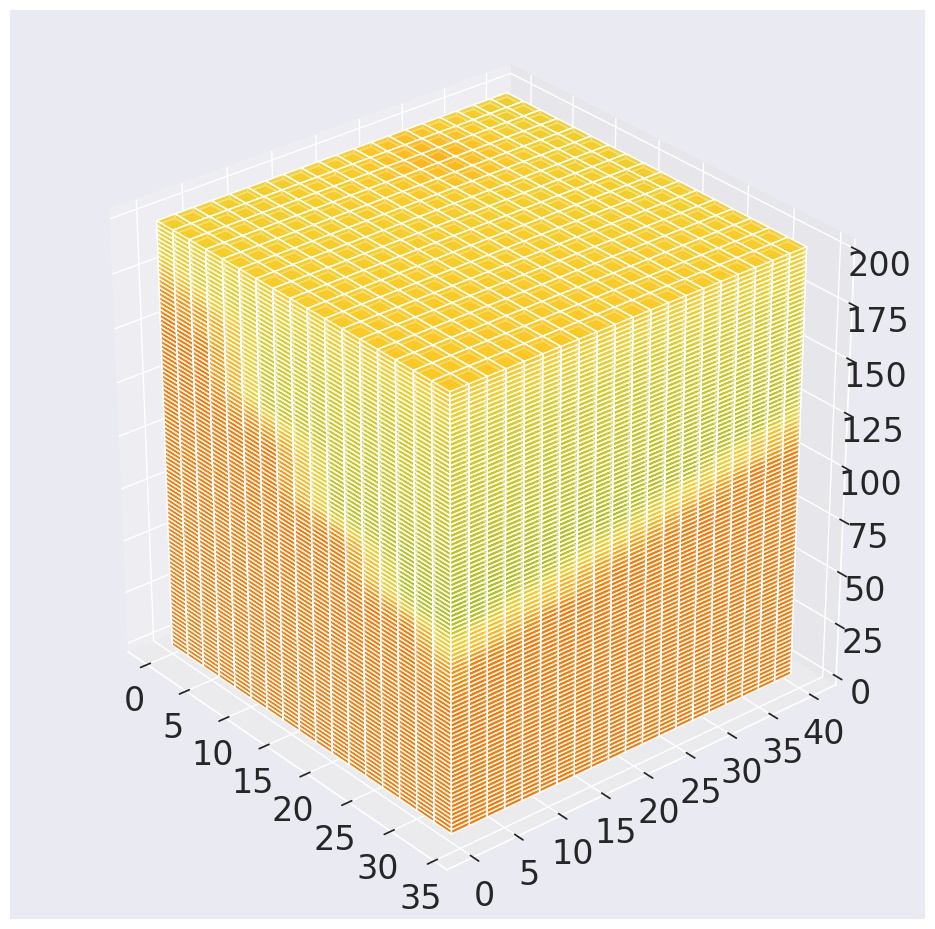}
\includegraphics[width=.21\textwidth]{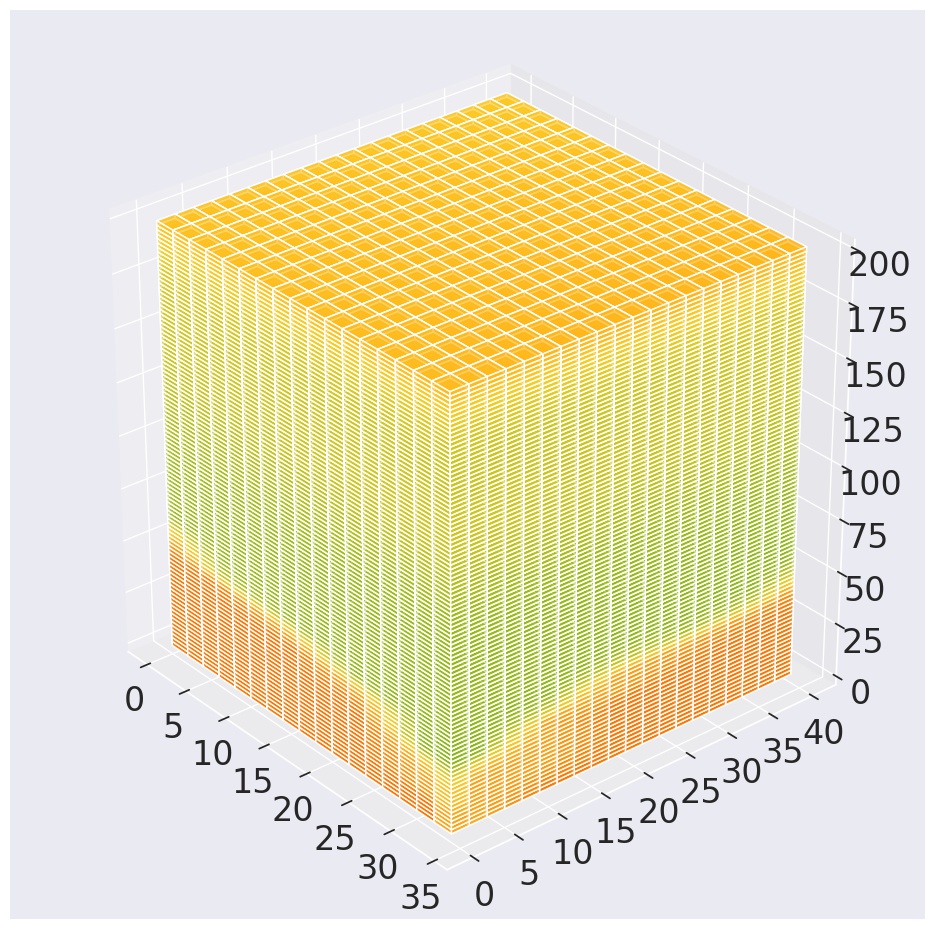}
\includegraphics[width=.21\textwidth]{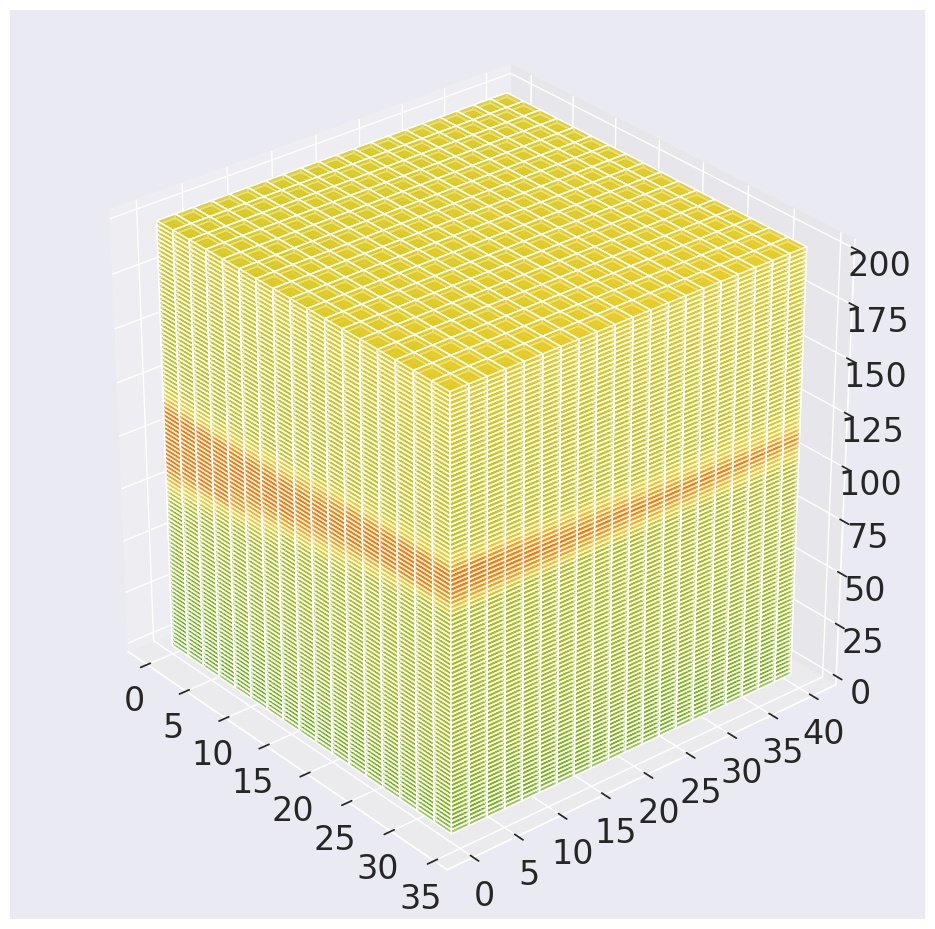}
    }

\mbox{\hspace{-0.2in}
 \includegraphics[width=.21\textwidth]{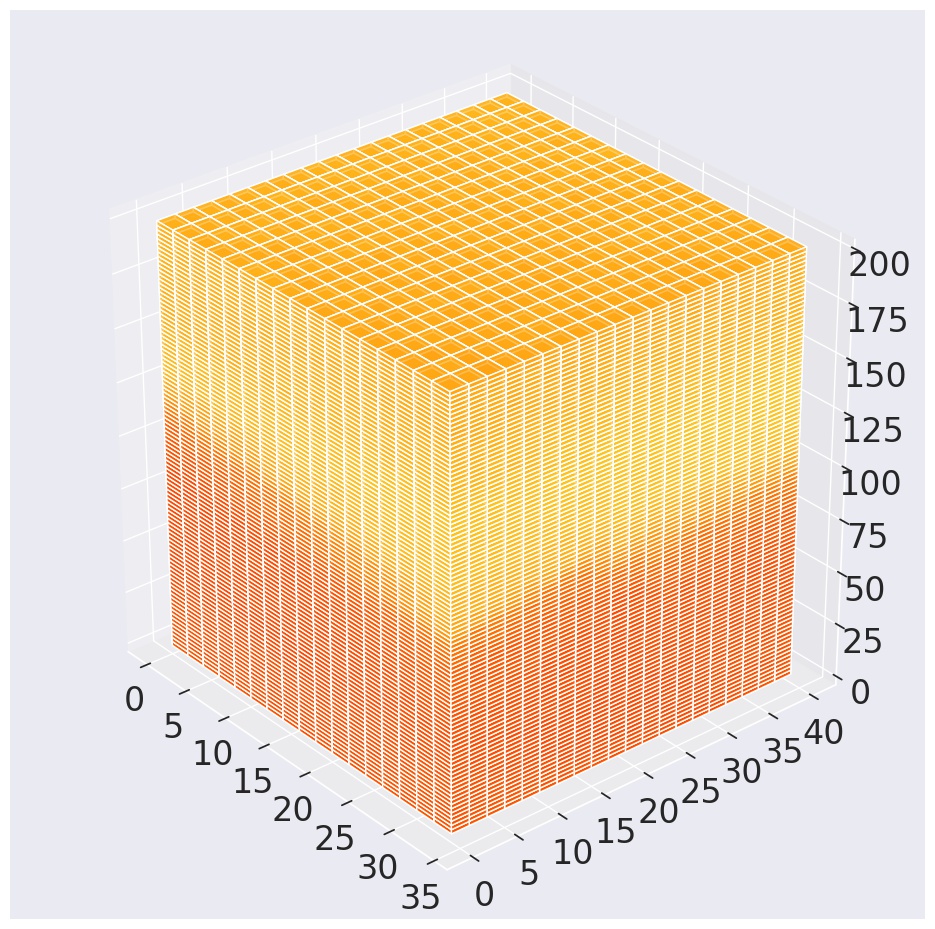}
\includegraphics[width=.21\textwidth]{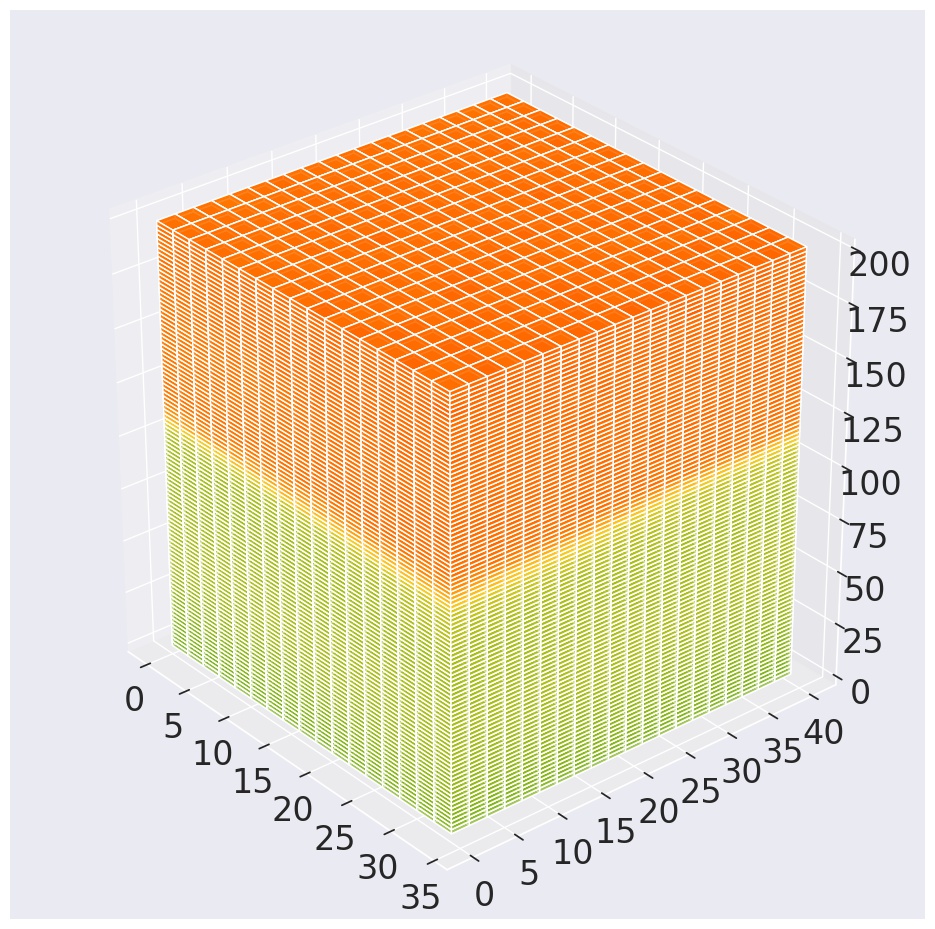}
\includegraphics[width=.21\textwidth]{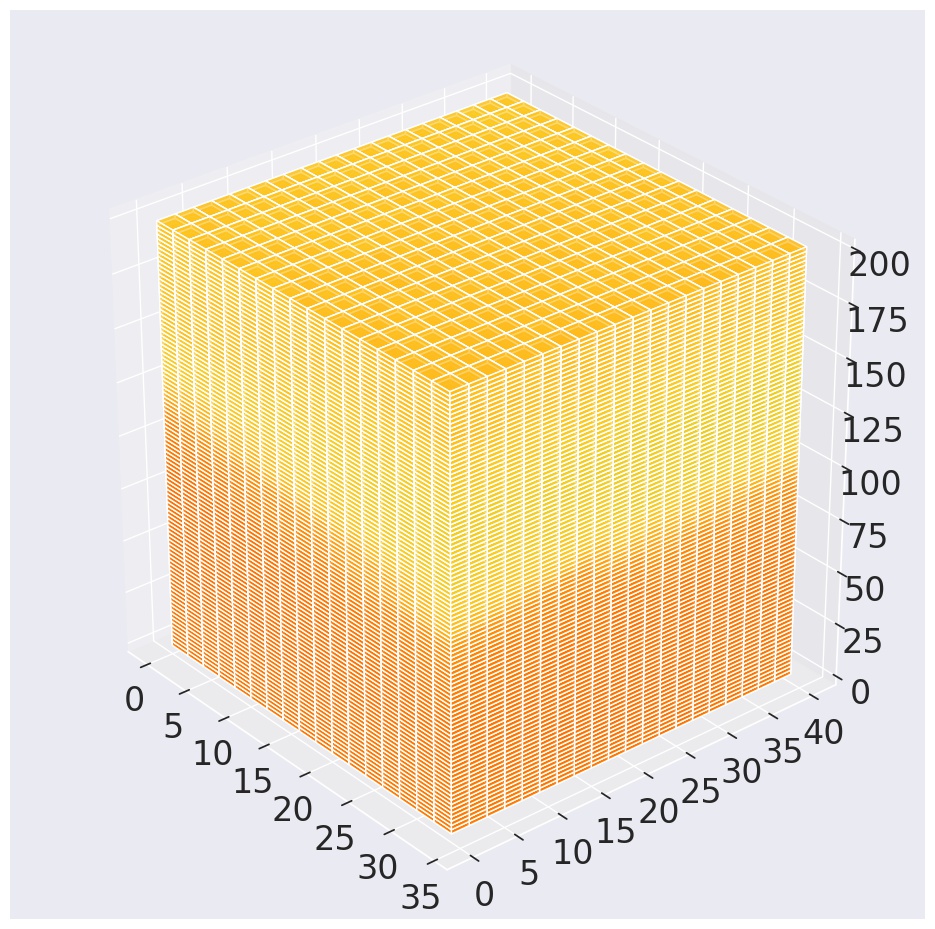}
\includegraphics[width=.21\textwidth]{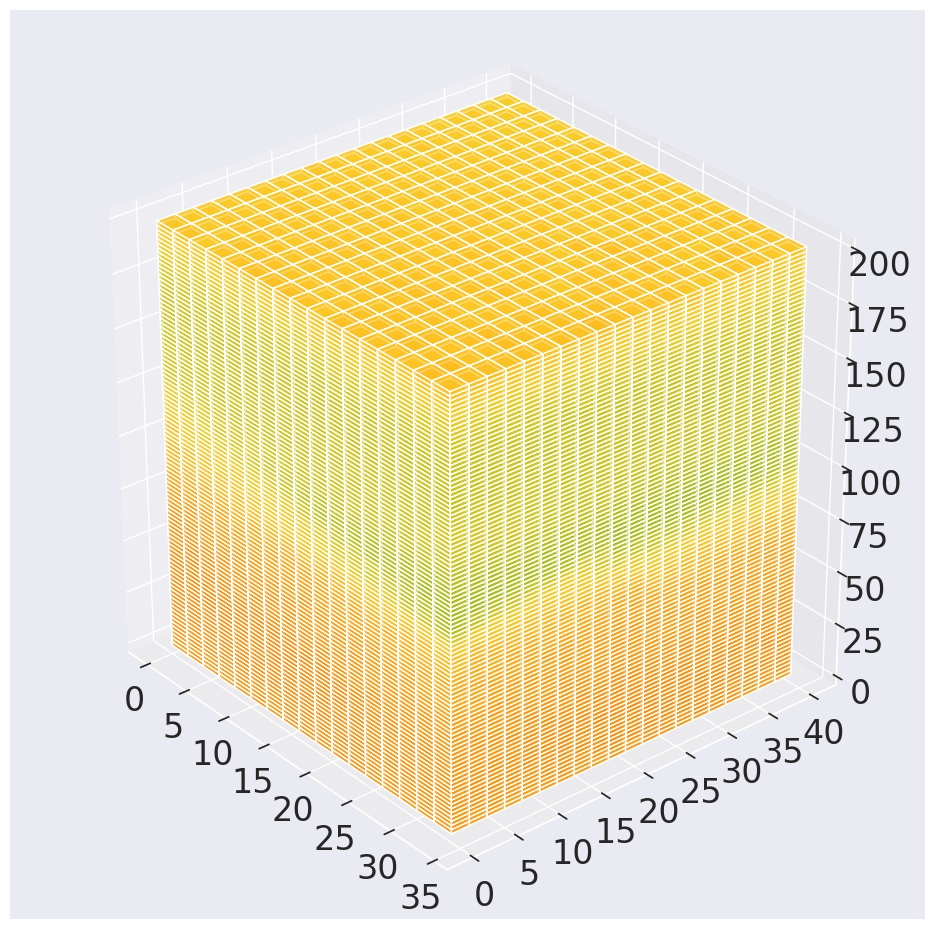}
\includegraphics[width=.21\textwidth]{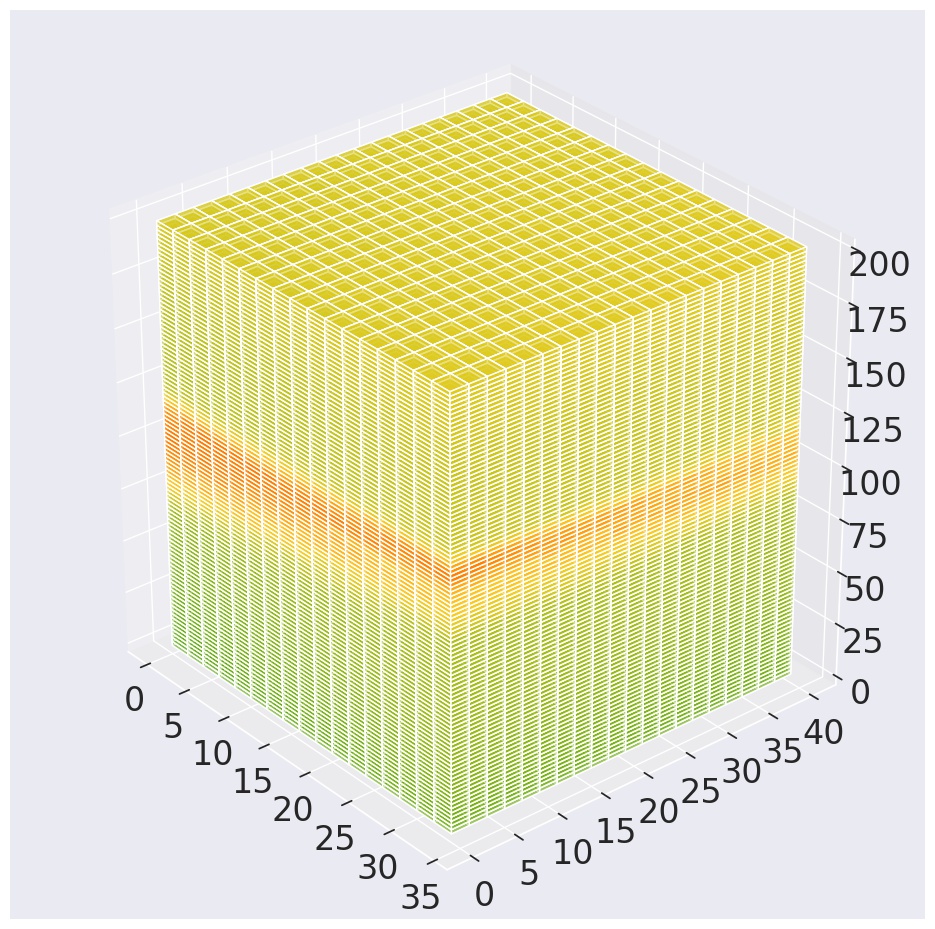}
    }
\mbox{\hspace{-0.2in}
\includegraphics[width=1.1\textwidth]{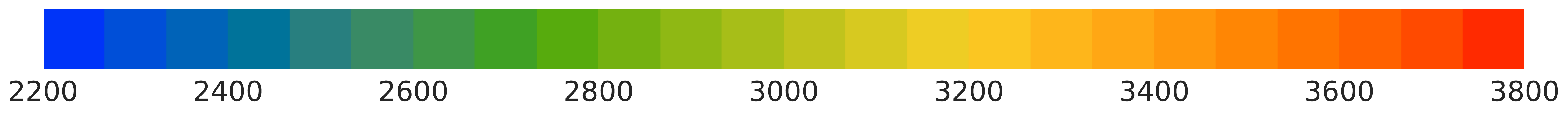}
}
\end{subfigure}
\hspace{-0.1in}
\begin{subfigure}{.35\textwidth}
\centering
\includegraphics[width=.8\textwidth]{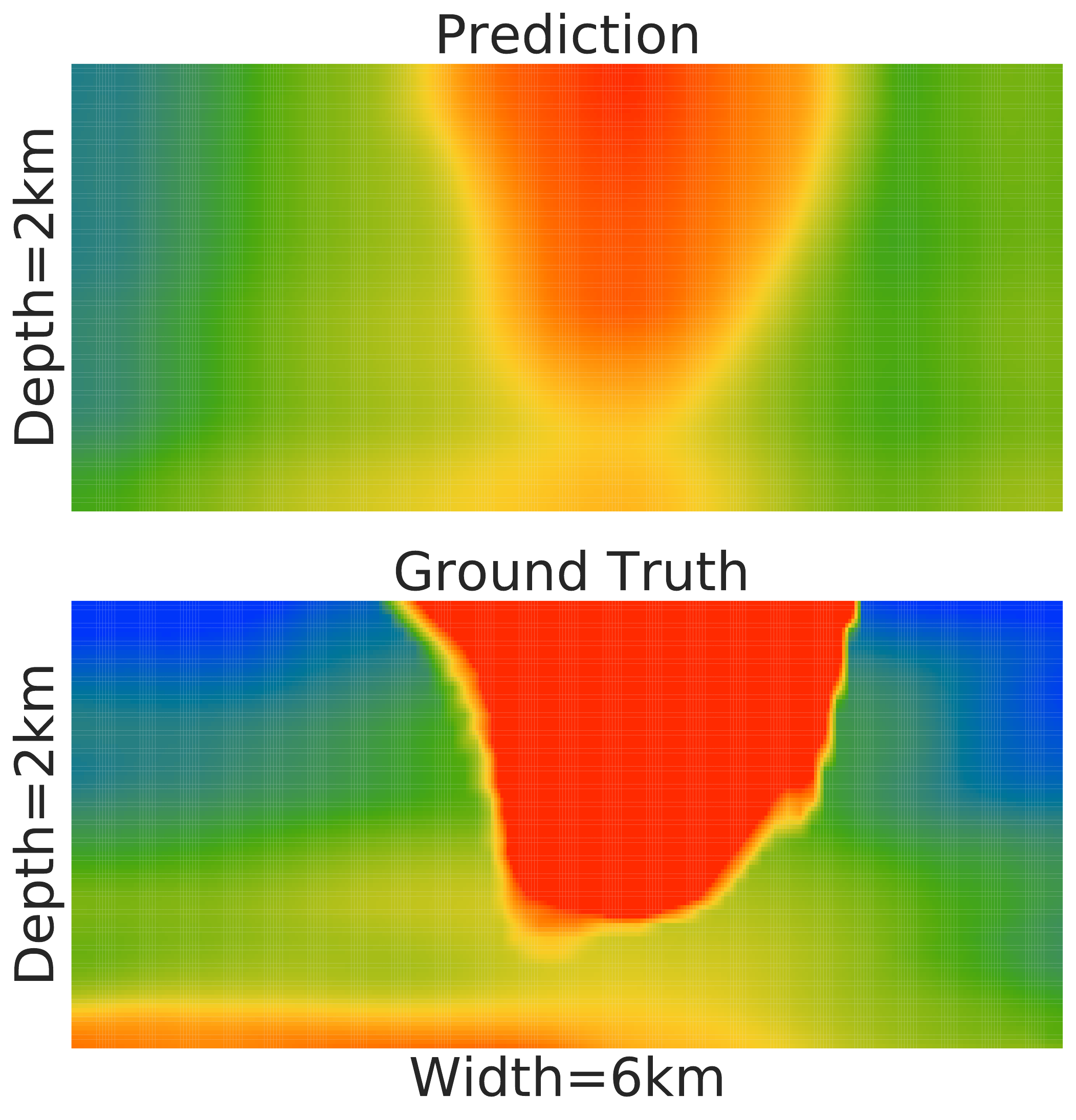}
\end{subfigure}
\caption{Visualization of prediction on GulfM-20 Dataset. \textbf{Left:} Top: ground-truth velocity cube, Bottom: predicted velocities. \textbf{Right:} cross-section visualization of inference on whole area. We want to stress that generating the above image from raw data via traditional methods takes several months because of the complex large scale data. Seismic data courtesy of TGS.}\label{fig:nakika100_demo}
\vspace{-0.6cm}
\end{figure*}
\subsubsection{Training Data} \label{sec:train_data}
% \vspace{-0.1cm}

($\mathcal{C}(q,w)$, $\mathcal{P}(q, w_p, d)$) is a training data sample for training a mapping $\mathcal{M}$ of depth corresponding to q. Given a huge survey area, we can sample different points q in the subsurface to generate multiple data points. Sometimes the availability of labeled real data is scarce. In such cases, we can sub-sample the context to generate more distinct training data points. Specifically, if $(\mathcal{C}, \mathcal{P})$ is a data point for training, then so is $(c, \mathcal{P})$ for every c in power set of $\mathcal{C}$, i.e. $c \in 2^{\mathcal{C}}$. We also observe that generalization of the model is better when trained with augmented subsampled data. Because SESDI is robust to different number of traces in the context, we can even use subsampling to improve the model's inference time in seismic processing.

% \vspace{-0.2cm}
\subsection{Benefits of SESDI}
% \vspace{-0.2cm}
\textbf{Large scale property model prediction:}
With the help of auxiliary task of contextual property model block, SESDI can efficiently predict a large scale property model.

\textbf{Robust to varying and irregular acquisition geometry:}
SESDI alleviates the issue of highly irregular acquisition and seamlessly works with varying amounts of data.

%Most CNN based networks require either regular acquisition geometry , listing of all possible source, reciever pairs \cite{Li_2020, Yang2019Deeplearning, inversionnet} or some feature extraction to convert the available data into regular data block \cite{semblance}. Regular Acquisition can almost never be achieved in real data gathering. Also, with reasonably sized 3D velocity blocks to predict, the area for relevant acquisition is considerably large. Number of all the possible (source, receiver) combinations even after reasonable grid quantization is very huge. These are exactly the reasons why it is difficult to apply models from current literature to real data.  Our model does not suffer from any such short-coming. Also, it does not require any kind of pre-processing and can work on raw traces. As is the case with real acquisitions, the number of traces in relevant acquisition area might vary from location to location. The model gracefully incorporates such requirements. 
\textbf{Scalable to k-dimensional real data}
Unlike current SDI approaches, it scales easily to 5-dimensional real data. SESDI, with its information-aligned trace embeddings can quickly ingest even higher dimensional seismic data obtained by adding more meta-details to each trace. 

%\textbf{Computationally scalable to real data}
%Most of the models in current literature work with 2D synthetic velocities where raw data takes a 3 dimensional form $(\text{source}_x, \text{receiver}_x, \text{time})$. Extending these models to 3D involves higher dimensional convolutions as now the data is at least 5 dimensional $(\text{source}_x, \text{source}_y, \text{receiver}_x, \text{receiver}_y, \text{time})$. These convolutions are very expensive to compute during both training and inference steps. SESDI, on the other hand , is not really affected by 2D / 3D inversion setups. The only change occurs in the dimension of trace-aq-info which has negligible effect on involved computation. Furthermore, there is even more meta-data associated with each trace. With convolutions it is impossible to fully utilize all the meta data due to increasing dimension of the convolutions. However, SESDI can easily be extended to find embeddings of this meta data and combining them into a trace-info embedding at minuscule additional cost. Hence SESDI has the potential to fully utilize entire available data.
\textbf{Does not overfit even under scarce data}
Labeled data in seismic tasks can be scarce. However, with the SESDI approach of generating data, we can create exponentially more distinct training samples from a small sample in data. (see \ref{sec:train_data}). Training model using this sampling approach prevents the model from overfitting.
For example, SEGSalt data has 130 training samples, but the SESDI model with sampling technique does not overfit even when we run the model for over 1200 epochs. (see figure \ref{fig:frac_overfit})
%Labelled data for the seismic tasks such as  VMB or SI is generally scarce. In order to train models, generally large amounts of synthetically generated data is used to train models. SESDI is especially resistant to over-fitting due to its architecture. Every sample of size say k for training SESDI can be essentially used to generate $2^k$ samples by subsampling the data. Hence, small number of samples can be used to train complicated SESDI model. We observe this even in our experiments, where the model test results keep on improving the longer we train without over-fitting . See Figure \ref{fig:frac_overfit} 

%\input{application}

%  \begin{table*}[]
%  \resizebox{\textwidth}{!}{
%  \begin{tabular}{|l|l|l|l|l|l|l|l|l|l|l|l|l|}
%  \hline
%  Datasets &
%   \multicolumn{3}{l|}{SEGSalt} &
%   \multicolumn{3}{l|}{RS-SEGSalt} &
%   \multicolumn{3}{l|}{GulfM-10} &
%   \multicolumn{3}{l|}{GulfM-20} \\ \hline
%  Models          & l1     & PSNR  & SSIM & l1     & PSNR  & SSIM & l1 & PSNR & SSIM & l1 & PSNR & SSIM \\ \hline
% U-Net           & 196.82 & 14.64 & 0.45 & -      & -     & -    & -  & -    & -    & -  & -    & -    \\ \hline
% Set-Transformer & 232.78 & 14.21 & 0.40 & 294.76 & 14.90 & 0.34 & -  & -    & -    & -  & -    & -    \\ \hline
%  SESDI &
%   \textbf{163.77} &
%   \textbf{16.65} &
%   \textbf{0.47} &
%   \textbf{162.79} &
%   \textbf{16.68} &
%   \textbf{0.41} &
%   135.27 &
%   32.34 &
%   0.82 &
%   164.99 &
%   26.95 &
%   0.75 \\ \hline
%  \end{tabular}
%  }
%  \caption{Summary of Results of SESDI and baselines on all the four datasets.}
%  \label{tab:all-results}
%  \end{table*}

  \begin{table*}[]
  \centering
 \resizebox{0.8\textwidth}{!}{
 \begin{tabular}{|l|l|l|l|l|l|l|l|l|}
 \hline
 Datasets &
   \multicolumn{3}{l|}{SEGSalt} &
   \multicolumn{3}{l|}{RS-SEGSalt} &
   \multicolumn{1}{l|}{GulfM-10} &
   \multicolumn{1}{l|}{GulfM-20} \\ \hline
 Models          & U-Net     & Set-Transformer  & SESDI & U-Net     &  Set-Transformer  & SESDI &  SESDI & SESDI  \\ \hline
$\mathcal{L}_1$          & 196.82 & 232.78 & \textbf{163.77} & -      & 294.76    & \textbf{162.79}    & 135.27  & 164.99      \\ \hline
PSNR & 14.64 & 14.21 & \textbf{16.65} & - & 14.90 & \textbf{16.68} & 32.34  & 26.95       \\ \hline
 SSIM &
   0.45 &
   0.40 &
   \textbf{0.47} &
   - &
   0.34 &
   \textbf{0.41} &
   0.82 &
   0.75 
   \\ \hline
 \end{tabular}
 }
 \vspace{-0.1cm}
 \caption{Summary of results of SESDI and baselines on all the four datasets.}
 \label{tab:all-results}
 \vspace{-0.5cm}
 \end{table*}
 % Second style table is in tab_all_res2.tex
 % that was very confusing. so removed

% \vspace{-0.2cm}
\section{Application and Empirical Results}
% \vspace{-0.2cm}
We evaluate the SESDI approach against the competing methods on the velocity inversion problem

% \vspace{-0.2cm}
\subsection{Baselines}
% \vspace{-0.2cm}
\begin{itemize}[leftmargin=*,nosep]
    \item \textbf{U-Net} \cite{Yang2019Deeplearning} is the state-of-the-art model in recent literature for data-driven seismic processing. Also, it is the only model that was evaluated on a publicly available SEGSalt data \footnote{https://wiki.seg.org/wiki/SEG}. Therefore, we choose this opensource  \cite{Yang2019Deeplearning} implementation as the representative for U-Net. 
    \item \textbf{Set-Transformer} \cite{lee2019set} is the state-of-the-art model that introduce multi-head attention mechanisms on set representation. It has demonstrated its superiority in point cloud classification and anomaly detection.   
    % \todo{@zhaozhuo add description. why you chose this baseline.}
\end{itemize}

The details on training strategies and hyper-parameters of these models can be found in supplement.

% \vspace{-0.2cm}
\subsection{Datasets}
% \vspace{-0.2cm}

To extensively evaluate SESDI against competing approaches, we carefully select the following datasets.

% \vspace{-0.2cm}
\begin{itemize}[leftmargin=*,nosep]
    \item \textbf{SEGSalt:} (small, regular, 2D Velocity). SEGSalt is a small scale, 2D velocity prediction dataset with regular acquisition (Fig.\ref{fig:syntheticAQ}). SEGSalt is an ideal dataset for the CNN-based U-net model used in \cite{Yang2019Deeplearning} and is publicly available. We choose this data set to compare against U-net at its best. \newline 
    \textbf{Data generation:} This data is generated by first designing 2D velocity models and then running forward wave propagation using tools like Devito \cite{devito-api}\cite{devito-compiler} to obtain seismic traces. The acquisition geometry is regular: distributing 29 sources and 301 receivers uniformly along the surface line with each receiver listening to each source. For details of data generation, refer to the supplement.

\item \textbf{RS-SEGSalt, Realistically-Sampled SEGSalt:} (small, irregular, 2D Velocity) RS-SEGSalt is obtained from SEGSalt by tweaking sampling the acquisition to make it more realistic. The way to sub-sample the traces is shown in figure \ref{fig:syntheticAQ}. With such a simple tweak towards real data, the U-Net model breaks and cannot be applied. \newline
\textbf{Data generation}: For each sample in SEGSalt for each epoch, we randomly sample a random fraction of contiguous traces. For experiments, we use a fraction uniformly sampled between $(0.7 \pm 0.25)$ and then choose the contiguous strips of receivers at a uniformly random location.
\item \textbf{GulfM-10 and GulfM-20}( real, enormous, highly irregular, 3D Velocity) These are raw, noisy datasets that we generate directly from real sensors obtained from a Big-Oil company without preprocessing steps like scaling, denoising, imputations, etc. This dataset contains missing values and highly irregular acquisition geometry, as shown in Figure \ref{fig:acquisition}. The data scale is also immense ( (100Km x 100Km) survey area). Due to the large scale, irregular geometry, this data set cannot be used by U-Net. We test set-transformer and SESDI on the auxiliary task of contextual velocity block prediction. We also show results on big block predictions by applying a sequence of contextual predictions and stitching together an image. \newline
\textbf{Data generation:} For the task of contextual prediction, we sample data points as described in \ref{sec:train_data}. The difference between GulfM-10 and GulfM-20 is the height of the target velocity block. GulfM-10 and GulfM-20 have velocity blocks with height 10\% (1Km) and 20\%(2Km) of the total velocity model, respectively. The block width and context width in both cases are 1Km and 5Km. 
\end{itemize}
% \vspace{-0.4cm}

\begin{figure}[]
\begin{center}
 \mbox{\hspace{-0.25in}
\includegraphics[width=0.97in]{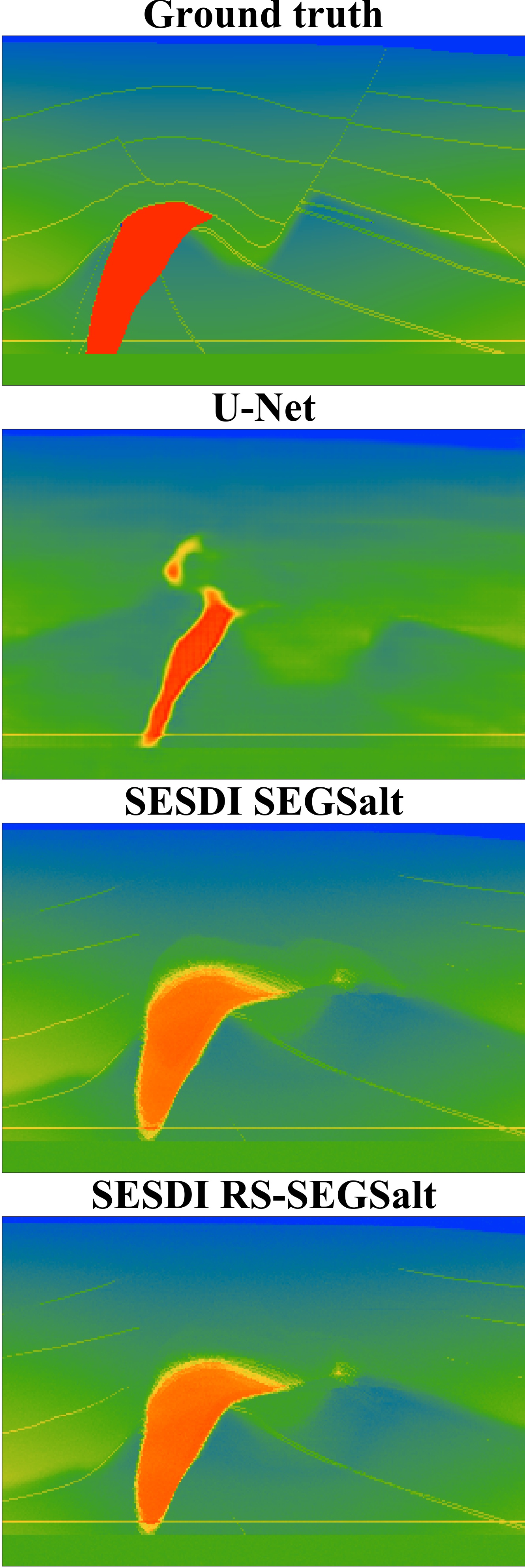}
\includegraphics[width=0.97in]{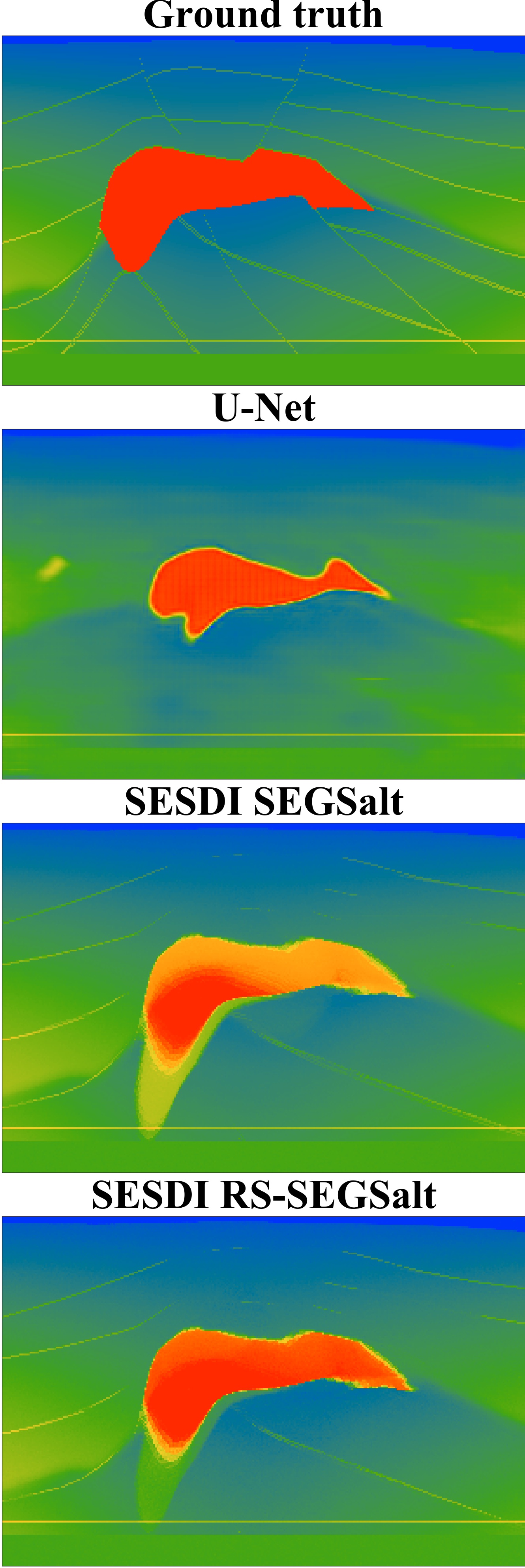}
\includegraphics[width=0.97in]{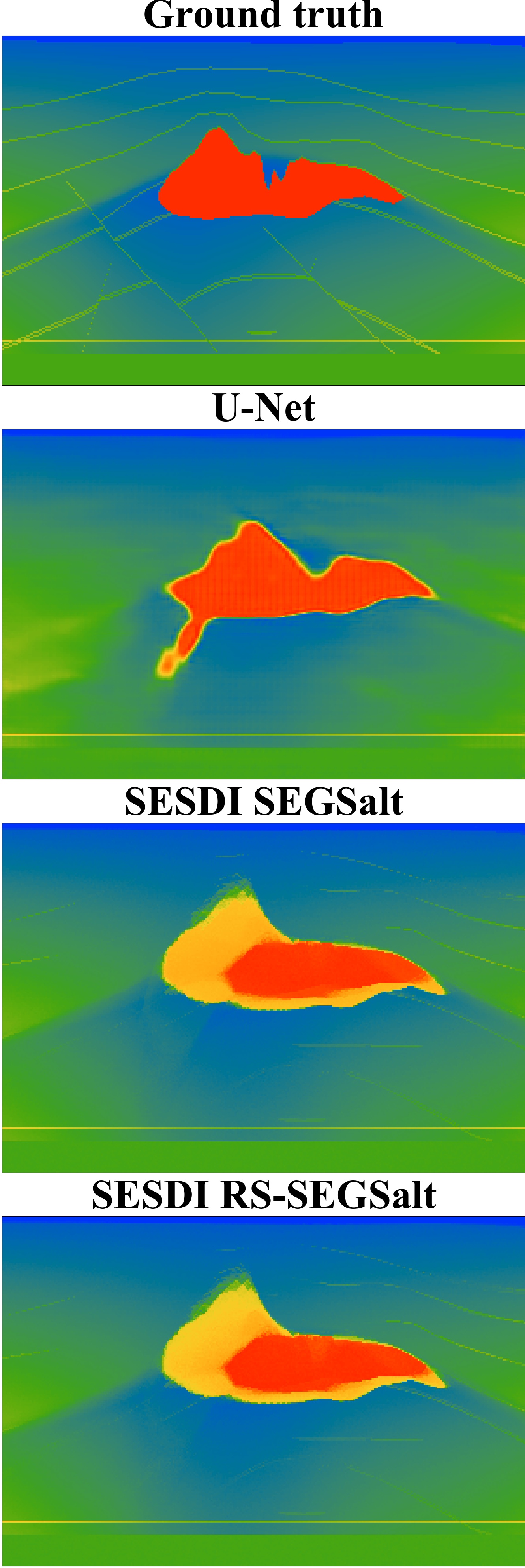}
    }

% \mbox{\hspace{-0.25in}

%     }

% \mbox{\hspace{-0.25in}

%     }
\mbox{\hspace{-0.25in}
\includegraphics[width=3in]{figures/rainbow/bar.pdf}
    }
\end{center}

\vspace{-0.15in}
\caption{Visualization of Prediction on SEGSalt Dataset : The prediction results for U-net vs SESDI model. SESDI performs particularly well when trained on RS-SEGSalt data. Also, SESDI model is robust to irregularities in the acquisition. }\label{fig:salt_demo}
\vspace{-0.6cm}
\end{figure}

\subsection{Metrics:}
% \vspace{-0.2cm}

To evaluate the models, we use the following three metrics widely used in recent work on data-driven seismic processing ~\cite{Yang2019Deeplearning,inversionnet,rojas2020physics,araya2018deep}. Namely, $\ell_1$ Loss (lower is better) , PSNR~\cite{huynh2008scope}(higher is better) and SSIM~\cite{wang2004image}(higher is better). $l1$ loss is $l_1$ metric applied to (input,output) images. SSIM is a structural similarity score widely used to measure the quality of images. PSNR (Peak Signal to noise ratio) is  used to measure image reconstruction quality.

% \vspace{-0.2cm}
\subsection{Results:}
% \vspace{-0.2cm}
%\begin{figure}
%\begin{center}
% \mbox{\hspace{-0.15in}
%
%\includegraphics[width=1.4in]{figures/fractionalsample/Test_l1.eps}
%\includegraphics[width=1.65in]{figures/fractionalsample/frac_l1.eps}
% \includegraphics[width=1.8in]{figures/overfitting/l1.png}
%}
% \mbox{\hspace{-0.15in}

%\includegraphics[width=1.4in]{figures/fractionalsample/Test_ssim.eps}
%\includegraphics[width=1.59in]{figures/fractionalsample/frac_ssim.eps}
% \includegraphics[width=1.8in]{figures/overfitting/l1.png}
%}
% \mbox{
%\hspace{-0.15in}
%\includegraphics[width=1.4in]{figures/fractionalsample/Test_PSNR.eps}
%\includegraphics[width=1.6in]{figures/fractionalsample/frac_PSNR.eps}
% \includegraphics[width=1.8in]{figures/overfitting/l1.png}
%}

%\end{center}

%\caption{\textbf{Left:} Evolution of metrics SSIM, L1 loss and PSNR on test SEGSalt dataset as the training progresses to 1200 epochs. As can be seen SSIM always improves and L1 loss , PSNR achieve stable values. For elaboration on metrics see Experiments section. Model is trained with 0.8 fraction of traces. \textbf{Right: }Comparison on test performance when different fractions of traces is used for inference}
%\label{fig:frac_overfit}

%\end{figure}

\begin{figure*}[]
    \centering % <-- added
\begin{subfigure}{0.32\textwidth}
  \includegraphics[trim=0 25 0 0 , clip, height=1.0in, width=0.99\linewidth]{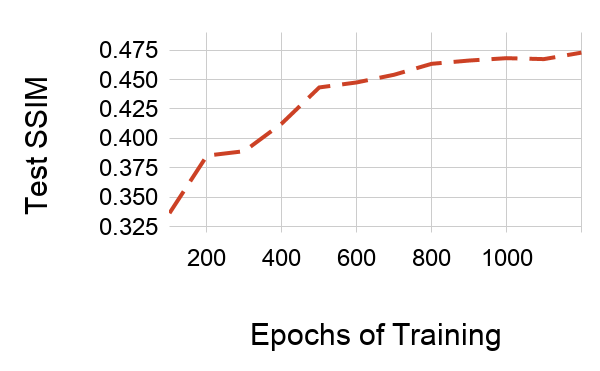}
  %\caption{image1}
  \label{fig:1}
\end{subfigure}
\begin{subfigure}{0.32\textwidth}
  \includegraphics[trim=0 25 0 0 , clip,height=1.0in, width=0.99\linewidth]{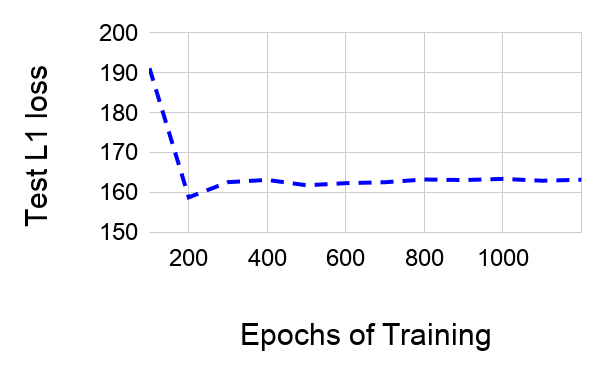}
  %\caption{image2}
  \label{fig:2}
\end{subfigure}
\begin{subfigure}{0.32\textwidth}
  \includegraphics[trim=0 25 0 0 , clip,height=1.0in, width=0.99\linewidth]{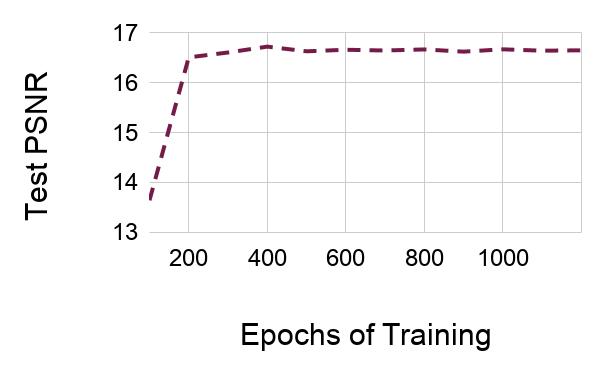}
  %\caption{image1}
  \label{fig:3}
\end{subfigure}

\begin{subfigure}{0.32\textwidth}
  \includegraphics[trim=0 25 0 0 , clip,height=1.0in, width=0.99\linewidth]{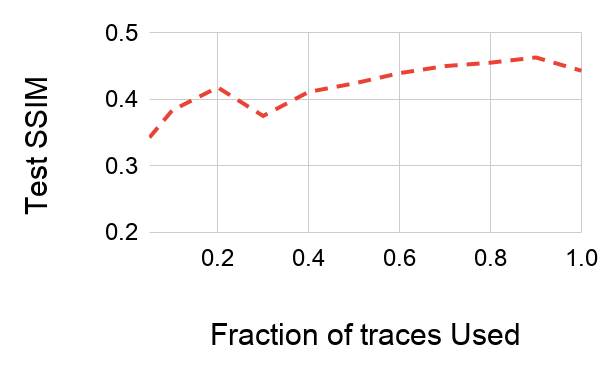}
  %\caption{image1}
  \label{fig:4}
\end{subfigure}
\begin{subfigure}{0.32\textwidth}
  \includegraphics[trim=0 25 0 0 , clip,height=1.0in, width=0.99\linewidth]{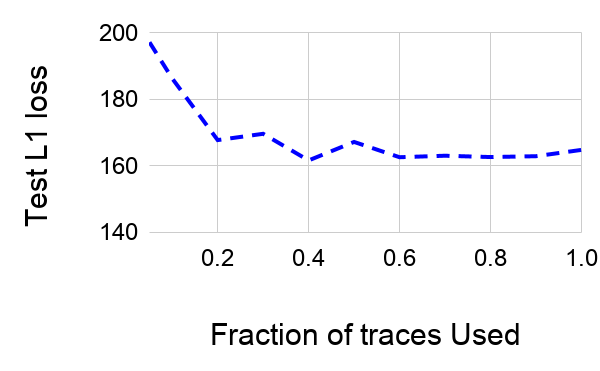}
  %\caption{image2}
  \label{fig:5}
\end{subfigure}
\begin{subfigure}{0.32\textwidth}
  \includegraphics[trim=0 25 0 0 , clip,height=1.0in, width=0.99\linewidth]{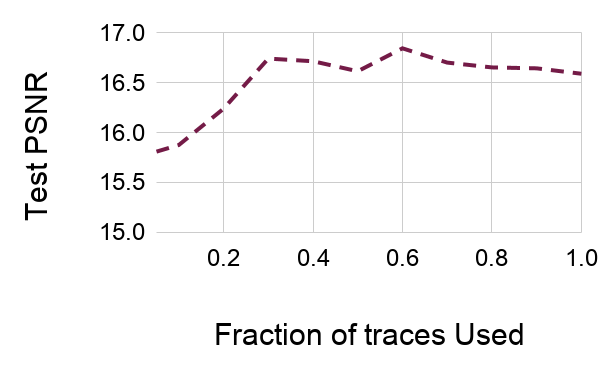}
  %\caption{image1}
  \label{fig:6}
\end{subfigure}
\vspace{-0.6cm}
\caption{\textbf{Above:} Evolution of metrics SSIM, $\mathcal{L}_1$  loss and PSNR on test SEGSalt dataset as the training progresses to 1200 epochs. As can be seen SSIM always improves and $\mathcal{L}_1$  loss , PSNR achieve stable values. For elaboration on metrics see Experiments section. Model is trained with 0.8 fraction of traces. \textbf{Below: }Comparison on test performance when different fractions of traces is used for inference}\label{fig:frac_overfit}
\vspace{-0.6cm}
\end{figure*}

We provide results in various forms:
% \vspace{-0.2cm}
\begin{itemize}[leftmargin=*,nosep]
    \item \textbf{Performance on Metrics (Table \ref{tab:all-results})} : lists the performance of all the models on all the datasets on all metrics.
    \item \textbf{Metric evolution and Robustness ( Fig. \ref{fig:frac_overfit})}: top panel shows the evolution of metrics. The bottom panel shows the performance of the SESDI model trained on RS-SEGSalt when used for inference on different sizes of the sampled data.
    \item \textbf{Visualization (Fig. \ref{fig:nakika100_demo}, \ref{fig:salt_demo} )}. Fig. \ref{fig:salt_demo} Shows the visualization of different models on SEGSalt. The entire test data visualization is presented in the supplement. We also show the performance of SESDI model trained on RS-SEGSalt data and inferred on SEGSalt. In Fig. \ref{fig:nakika100_demo}, we show some visualisation on contextual block prediction task (left). Also, we run inference on larger blocks of velocity and show visualization for the same (right). Details of visualization are presented in the appendix. \newline
\end{itemize}
% \vspace{-0.3cm}
\textbf{Key Highlights:}
% \vspace{-0.1cm}
\begin{itemize}[leftmargin=*,nosep]
    \item \textbf{U-Net vs. SESDI On SEGSalt:} On SEGSalt, which is the ideal data set for CNN based U-Net, SESDI performs better than U-net on all the metrics with significant margin (Table \ref{tab:all-results}). This begs the question of whether convolutions are apt for seismic inversion. Convolution based networks are known for their strong priors and disregard for long-range dependencies. MLPs, on the other hand, can capture all range dependencies and hence perform better. Visualizations in figure \ref{fig:salt_demo} also echo similar results. At times, U-Net incorrectly predicts background velocity and shows weird salt structures. Though SESDI's output suffers from over-smoothening the boundaries, it captures the salt and background's general shape in detail.
    \item \textbf{RS-SEGSalt vs. SEGSalt} Training the model on RS-SEGSalt leads to robust models due to the availability of extensive, varied data created by subsampling. Note that RS-SEGSalt only differs from SEGSalt in the input. Thus, we can train the model on RS-SEGSalt and test it on SEGSalt. We can see from figure \ref{fig:salt_demo} that SESDI gives better salt shapes when trained on RS-SEGSalt. This is also reflected by better $\mathcal{L}_1$  and PSNR for SESDI.
    \item \textbf{SESDI vs Set-Transformer} Table \ref{tab:all-results} also demonstrates the superiority of SESDI over set-transformer. On regular observation, SESDI achieves over 20$\%$ performance boost on set-transformer. One intuition behind this results is that  there is less dependency between traces. Thus, attending one trace with others would not improve the generalization of the set model.
    
    % \todo{@zhaozhuo this is very cryptic. you should elaborate. removed other points as they are already present in above points}
    
    \item \textbf{SESDI on GulfM}.  Figure \ref{fig:nakika100_demo} (right), shows that we get geologically sound predictions of the velocity model by using SESDI model. The image is obtained by stitching together the individual blocks predicted by SESDI model. Figure \ref{fig:nakika100_demo} (left) shows a mixture of small block predictions. 
    %We want to stress that obtaining an image like that shown above from unprocessed raw data takes traditional methods several months.
    SSIM on real velocity prediction is higher than that on synthetic data.  Real velocities in a small block are relatively smooth, whereas synthetic data is filled with complexities to resemble the entire survey.
    
\end{itemize}

%\begin{table}[]
    %\resizebox{\linewidth}{!} {
%\begin{tabular}{|l|l|l|l|l|}
%\hline
%                & SEGSalt & RS-SEGSalt & GulfM-10 & GulfM-20 \\ \hline
%U-Net           & Y       & N          & N        & N        \\ \hline
%Set-transformer & Y       & Y          & N        & N        \\ \hline
%SESDI           & Y       & Y          & Y        & Y        \\ \hline
%\end{tabular}
%}
%\caption{U-Net cannot be run on other datasets due to either irregular geometry, large scale or both. Set-transformer performs bad on RS-SEGSalt and hence we do not invest in experiments on GulfM datasets. SESDI is run on all datasets.}

%\label{tab:baseline-dataset}
%\end{table}

%Not sure what the difference is between Discussion and conclusion .SO COMMENTING OUT.
\vspace{-0.2cm}
\section{Conclusion} 
\vspace{-0.2cm}
To tackle the scale of seismic inversion , we introduce an auxiliary task of contextual small property model block prediction to solve the problem of large-scale property model (in Gigabytes) prediction from Terabytes of data. We can efficiently train models for this auxiliary task. We alleviate the issues of highly irregular geometry of acquisition by formulating the Context as a bag of informative traces and provide a SESDI model to align and combine this data. Our model achieves geologically sound predictions on real data at its original scale. We believe this is the first-ever demonstration of a model on real scale seismic data.

\textbf{Acknowledgement:} The authors thank Shell International E$\&$P for financial support, computational resources, and permission to publish this work. The authors also thank TGS for providing the data.

% \input{experiments}

\newpage
%\blue{
%\subsection{Notation recommendation}
%\begin{tabular}{l|l}
%\hline
%$t$ & geometry aware trace \\
%$\mathbb{AG}$ & Complete Acquisition geometry\\
%$\mathcal{A}(t)$ & Acquisition geometry of a particular trace t\\
%$\mathcal{D}(t)$ & data of a particular trace t\\
%$\mathcal{P}(q,w_p,d)$ & proper model block with mid point at \\
%& location q, and has dimensions $w_p \times w_p \times d$\\
%$\mathcal{C}(q, w)$ & context at q.x,q.y of width $w\times w$\\
%\hline
%\end{tabular}
%}
\bibliography{ref}
\bibliographystyle{icml2021}
\newpage
\onecolumn
\appendix
\section{Experiment Details}\label{apx:exp}
\subsection{Seismic Data Simulation}\label{apx:sim}

The 2D SEG salt models in SEGSalt dataset~\cite{Yang2019Deeplearning} are identical to each other. The data statistics are shown in Table~\ref{apx:stats}. \cite{Yang2019Deeplearning} assume that the velocities varies from 2000 m/s to 4500 m/s, where the material with 4500 m/s is the salt body~\cite{onajite2013seismic}.  The shape of the target velocity matrix is 201$\times$301 with a spatial
interval of 10m.

The simulation parameters for trace generation are shown in Table~\ref{param:sim}. We perform a the time-domain
stagger-grid finite-difference scheme to simulate traces by acoustic wave equation.  The scheme contains both a second-order time direction and eighthorder space direction~\cite{ozdenvar1997algorithms}. For acquisition geometry, we assume there exists 301 receivers that are uniformly  distributed at a constant spatial interval. There also exists a perfectly matched
layer (PML) ~\cite{komatitsch2003perfectly} absorbing boundary condition that reduces unphysical reflection on edges. Therefore, the trace is a $2000$ dimensional vector and each velocity matrix corresponds to $29\times 301$ traces. To fit the requirements of the CNN, each trace is down-sampled to $400$ dimension by~\cite{Yang2019Deeplearning}. Then, the $301$ traces of each shot formulate a $301\times 400$ matrix. Therefore, the U-Net operations can performed on $ 301\times 400\times29$ image. Our SESDI model formulate each sample as a set with size $n<29\times301$  and element dimensionality $400$. We sub-sample the $29\times301$ traces to generate set embedding and predict the target velocity matrix.

\begin{table}[h]
\begin{center}
\setlength{\abovecaptionskip}{3pt}
\setlength{\belowcaptionskip}{2pt}
 \caption{Data Statistics for 2D VMB}
 \begin{tabular}{ p{2cm}|p{1.2cm}|p{1.2cm}|p{1.5cm} }
 \hline
 Dataset &  Training Samples & Testing Samples & Velocity Shape  \\
 \hline\hline
% Simulate & 1600 & 100 & 201$\times$301  \\
SEGSalt & 130 & 10 & 201$\times$301  \\
GulfM-10 & 400 & 100 & 17$\times$20$\times$50  \\
GulfM-20  & 400 & 100 & 17$\times$20$\times$100  \\
 \hline
 
\end{tabular}\label{apx:stats}
\end{center}
\end{table}

\begin{table*}[h]
\begin{center}
\setlength{\abovecaptionskip}{3pt}
\setlength{\belowcaptionskip}{2pt}
 \caption{Simulation Parameters}
 \begin{tabular}{ p{1.5cm}|p{1.5cm}|p{1.5cm}|p{1cm}|p{1.5cm}|p{1.5cm} }
 \hline
 Number of Source &   Number of Receivers & Sampling Frequency & Ricker Wave & Simulation Time & Traces Length \\
 \hline\hline
29 & 301 & 1kHz & 25Hz & 2s & 2000  \\
 \hline
 
\end{tabular}\label{param:sim}
\end{center}
\end{table*}

\subsection{Network Architecture}\label{apx:net}

We present the SESDI network architecture on Table~\ref{apx:net1}. The trace input is first embedded as a $10240$ dimension vector and then pass through a 3 layer MLP. The location information is embedded as a 512 dimension vector by a 4 layer MLP. Then, the location embedding is concatenated with the trace hidden vector and all the $n$ trace-location vectors are summed together. Finally, a hidden layer and an output layer are used to generate the final velocity matrix. All the MLP in this model are associated with ReLU activation. For set transformer~\cite{lee2019set}, we implement an encoder with a stack
of set attention blocks. Then, we introduce a pooling by
multihead attention module. Finally, we have a stack of set attention blocks in decoder. We set the dimensionality of all hidden layers to 4096 and the number of attention heads to 4. 

\begin{table*}[h]
\begin{center}
\setlength{\abovecaptionskip}{3pt}
\setlength{\belowcaptionskip}{2pt}
 \caption{SESDI Network Architecture 2D VMB}
 \begin{tabular}{ p{2.5cm}|p{2cm}|p{2cm} |p{2cm}}
 \hline
 Layer &  Type & Input Dim & Hidden Units   \\
 \hline\hline
Trace Emb & 1 Layer MLP & 400 & 10240  \\
Trace Hidden & 3 Layer MLP & 10240 & 4096  \\
Location Emb & 4 Layer MLP & 2 & 512  \\
Hidden & 5 Layer MLP & 4096+512 & 4096  \\
Output & 1 Layer MLP & 4096 & 201$\times$301  \\
 \hline
 
\end{tabular}\label{apx:net1}
\end{center}
\end{table*}

\subsection{GulfM Dataset}\label{apx:nakika}

The GulfM dataset is generated from the real marine survey data shared by one of the major energy and petrochemical company. As shown in the name, the raw seismic data is obtained by oil and gas platform~\cite{crout2008oil} in the gulf of mexico. The large ground-truth 3D velocity models are generated by geophysicists with duration over 8 months. We focus on the velocity cubes in $100m$ to $200m$ below the surface and random sample cubes within $33 000m\times78000m$ area. In other words, we random sample velocity $1000m\times1000m\times50m$ and $1000m\times1000m\times100$ cubes from a large $33 000m\times78000m\times 100m$ cubes to generate labels for GulfM-10 and GulfM-20 dataset. After discretization, the output velocity shapes for two GulfM datasets are $17m\times20m\times50m$ and $17m\times20m\times100$. Then, for each velocity cubes, we acquire its corresponding traces collected on the $1000m\times1000m$ surface area as SESDI model's input. Because the complex acquisition geometry, the maximum number of receivers for each trace is 7008. And the each velocity cube has traces varies from $2$ to $8779$. This set format prohibit convolution models in this task.  

\subsection{Model Training}\label{apx:train}

We train both SESDI model and U-Net on a server with 1 Nvidia Tesla V100 GPU and two 20-core/40-thread processors (Intel Xeon(R) E5-2698 v4 2.20GHz). SESDI, set transformer and U-Net use Adam as optimizer. The training hyper-parameters for SESDI, U-Net and set transformer follow the FCNVMB paper~\cite{Yang2019Deeplearning} with learning rate modifications to make better performance.
\begin{figure*}[htb]
\begin{center}
 \mbox{\hspace{-0.2in}
 \includegraphics[width=.15\textwidth]{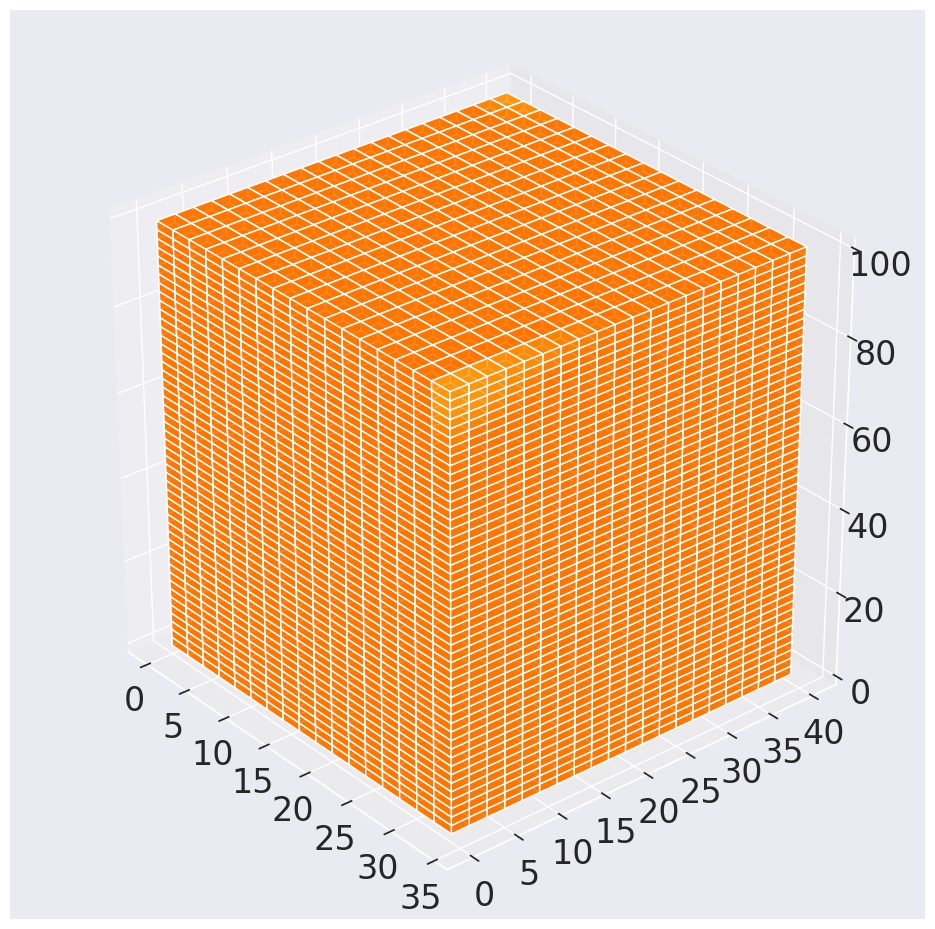}
\includegraphics[width=.15\textwidth]{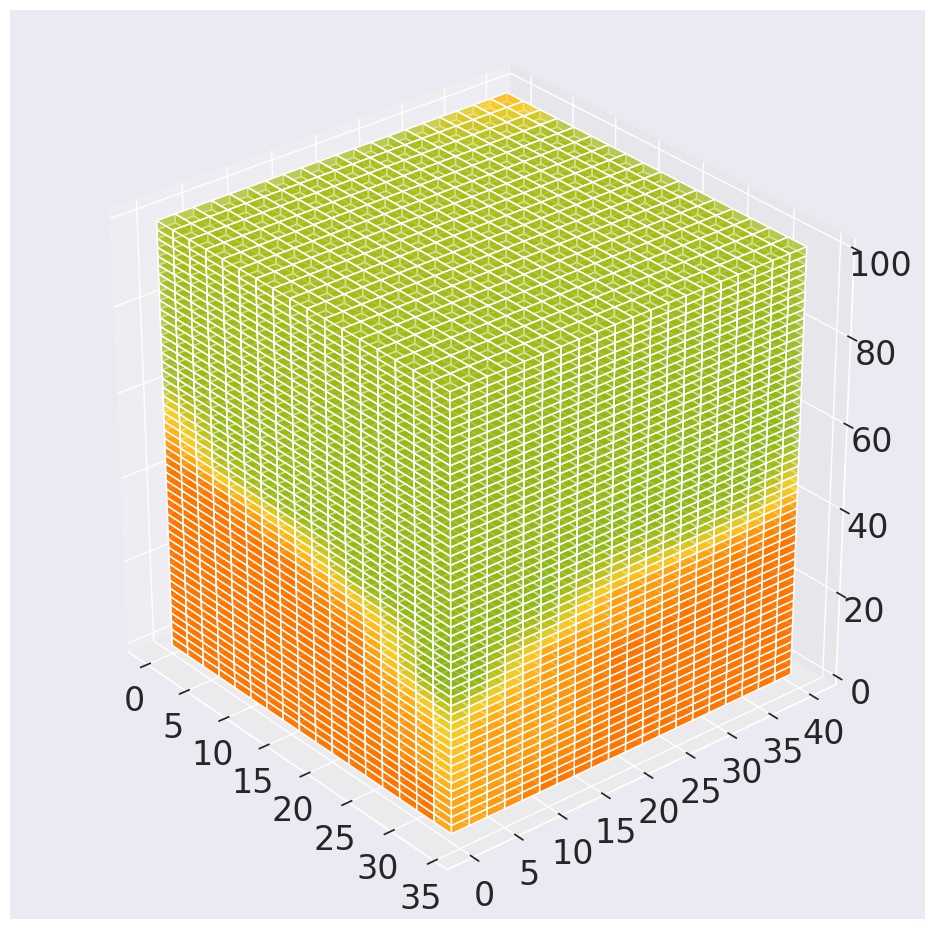}
\includegraphics[width=.15\textwidth]{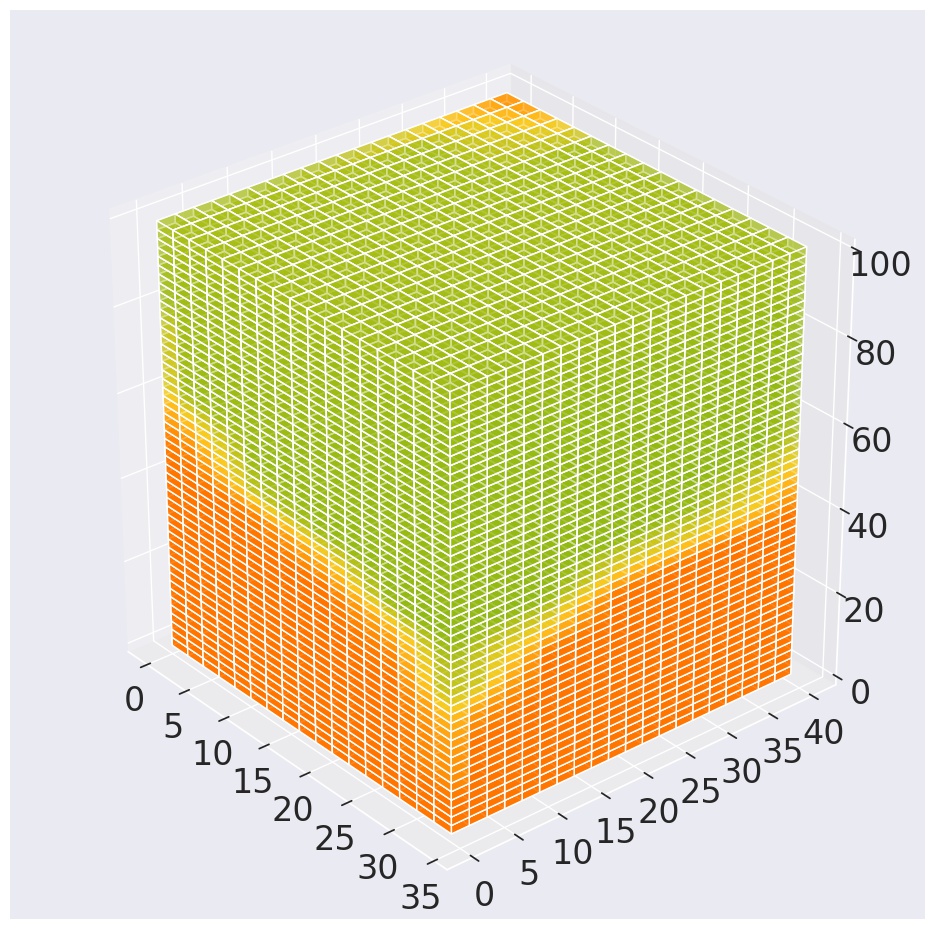}
\includegraphics[width=.15\textwidth]{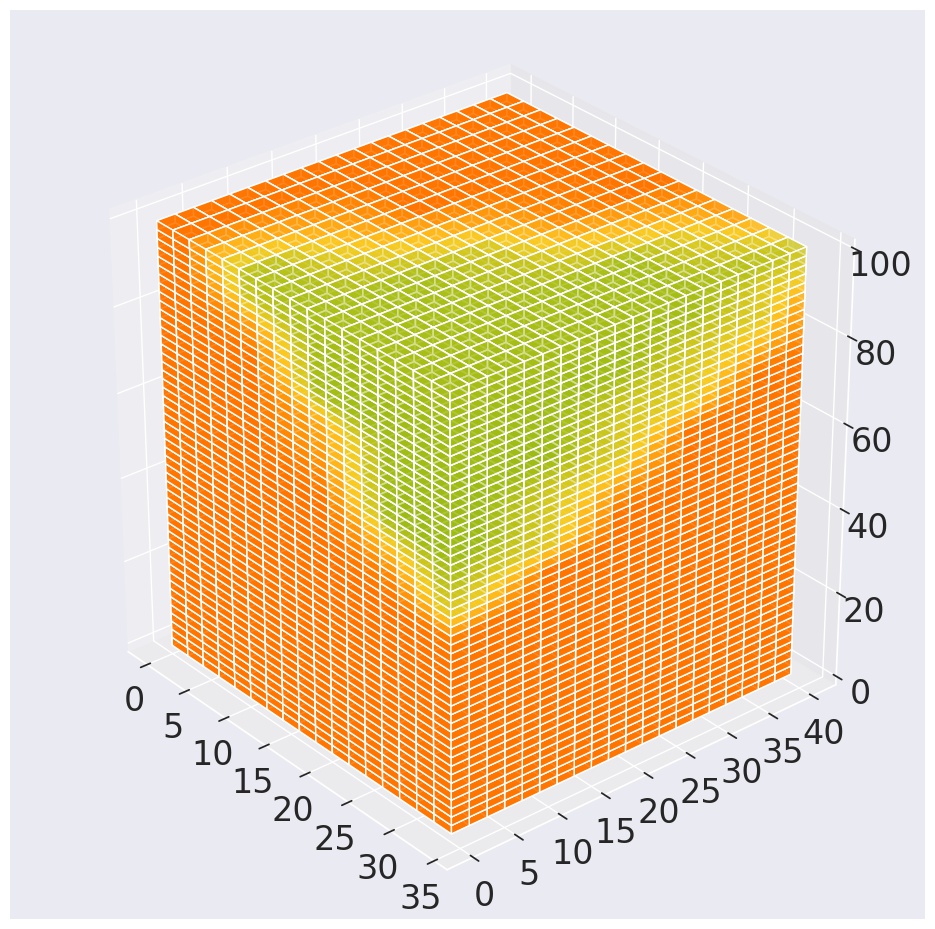}
\includegraphics[width=.15\textwidth]{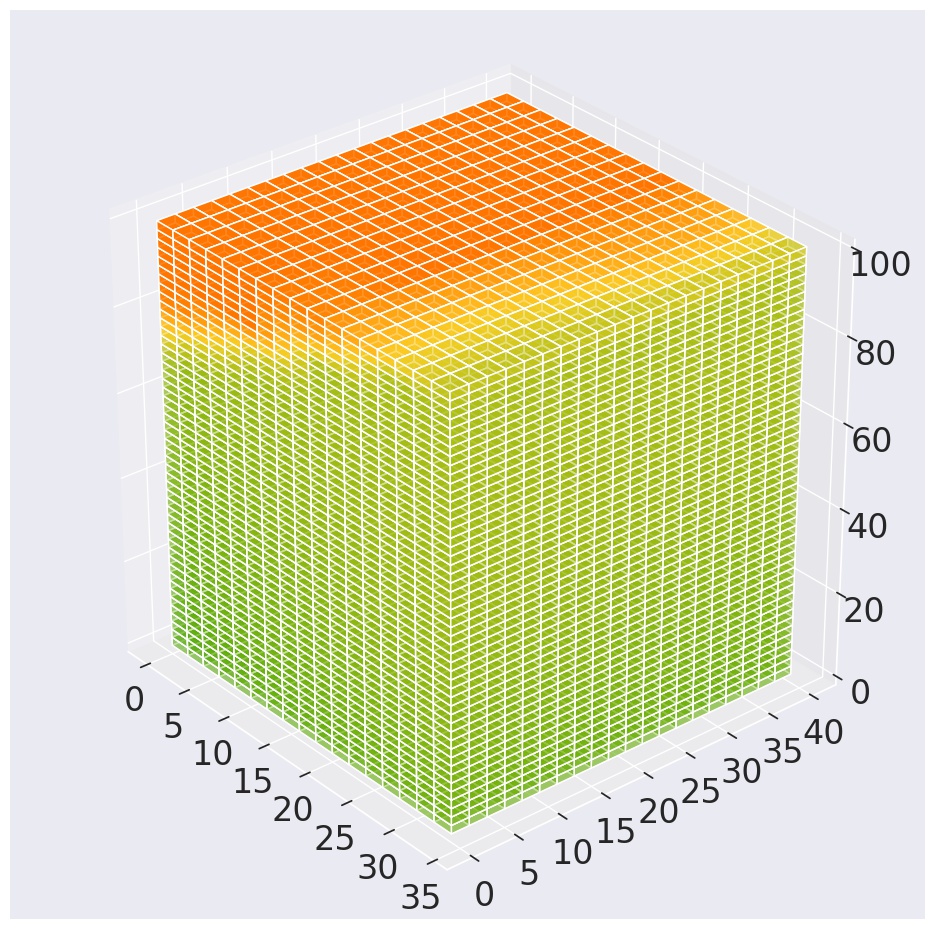}
    }

\mbox{\hspace{-0.2in}
 \includegraphics[width=.15\textwidth]{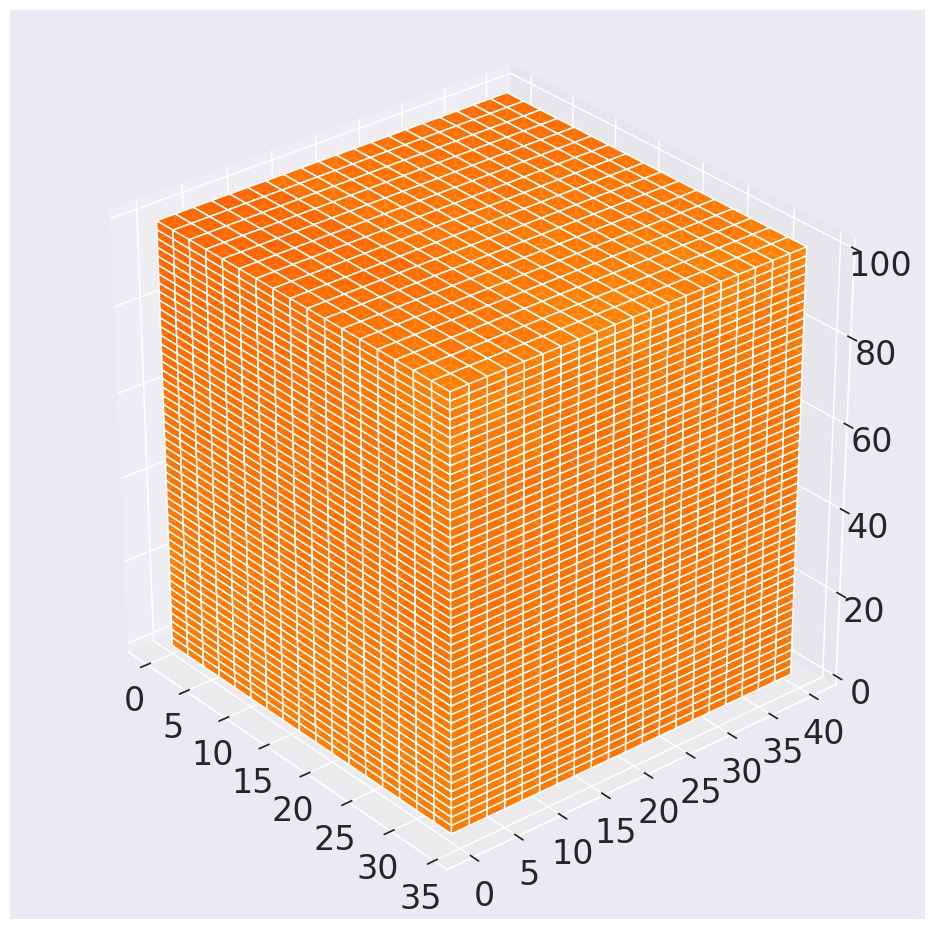}
\includegraphics[width=.15\textwidth]{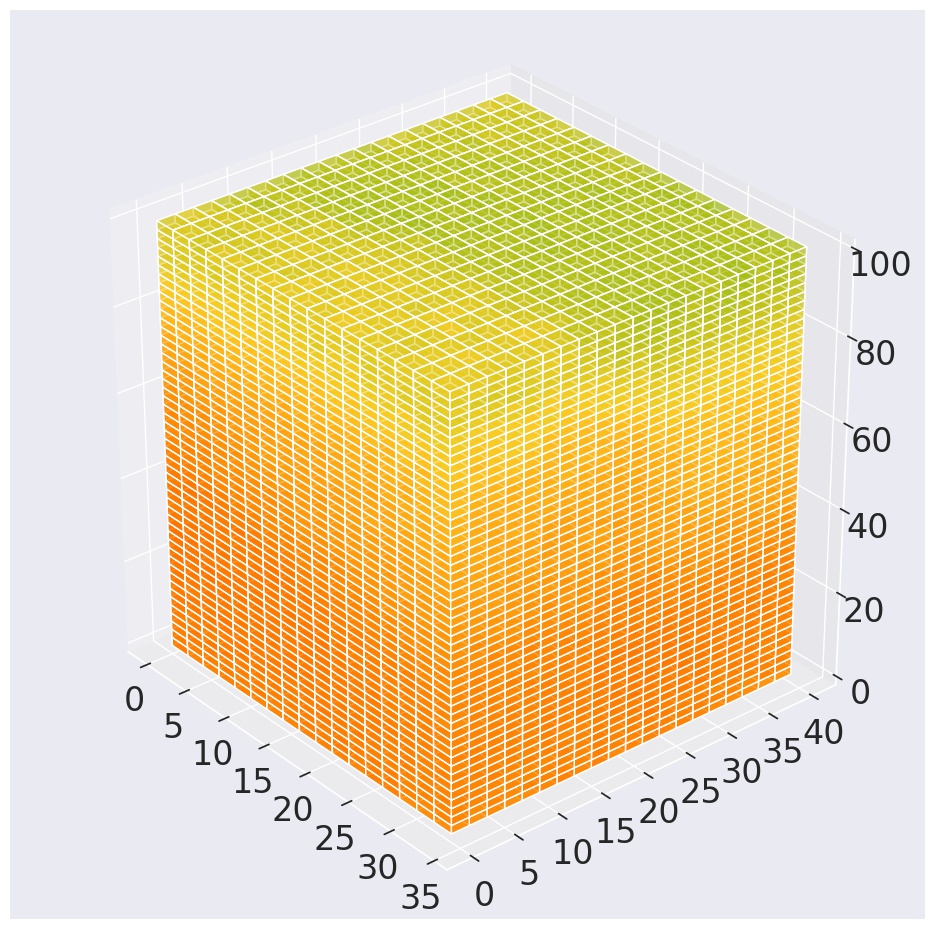}
\includegraphics[width=.15\textwidth]{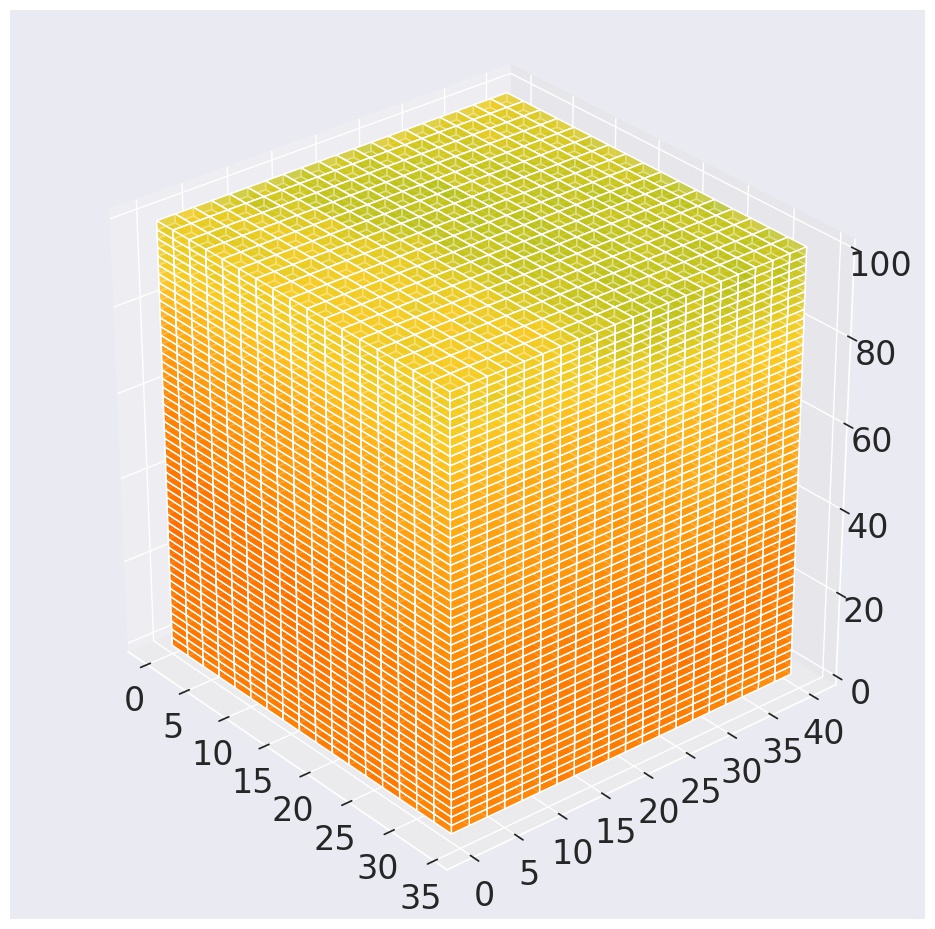}
\includegraphics[width=.15\textwidth]{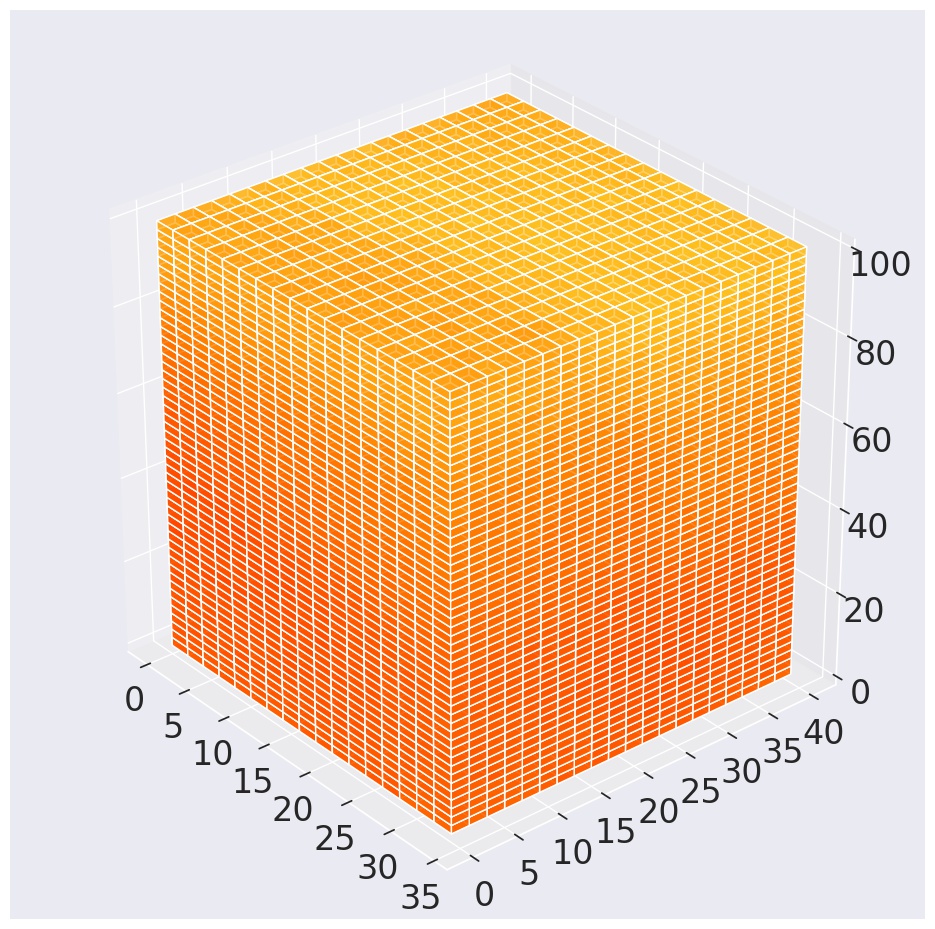}
\includegraphics[width=.15\textwidth]{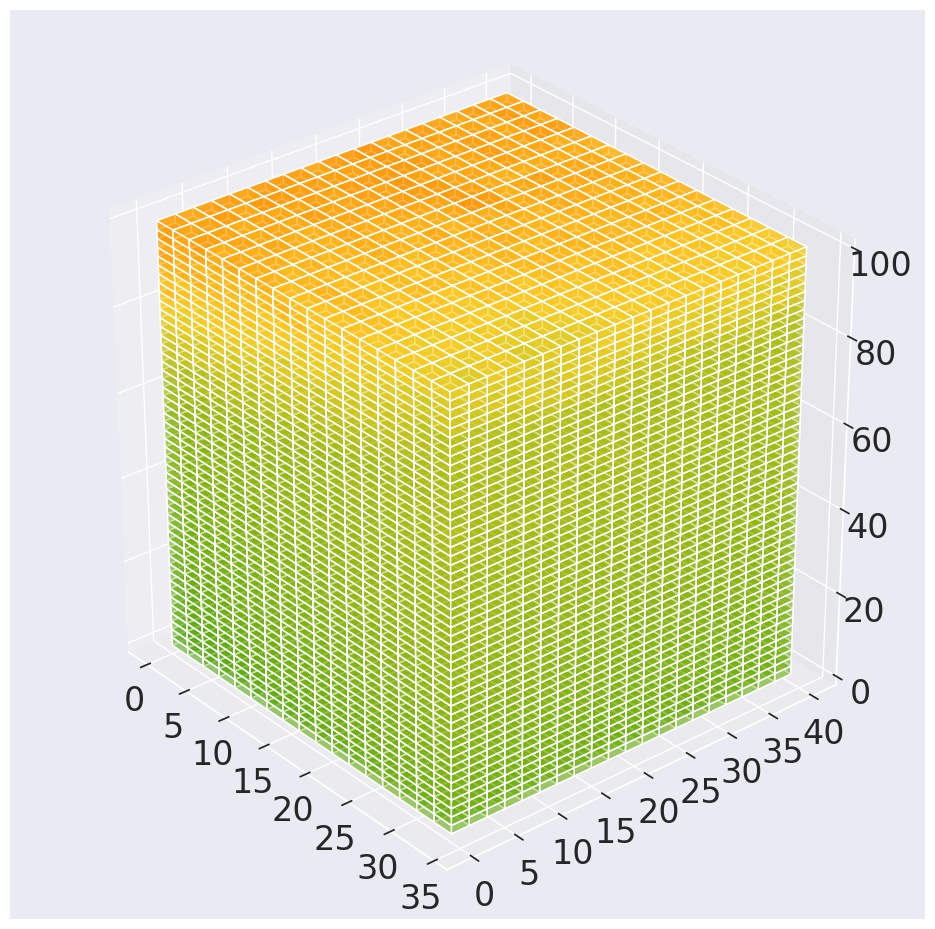}
    }
\mbox{\hspace{-0.2in}
\includegraphics[width=4.5in]{figures/rainbow/bar.pdf}
}

\end{center}

\vspace{-0.15in}
\caption{Visualization of Prediction on GulfM-10 Dataset. Seismic data courtesy of TGS.}\label{fig:nakika50_apx}
\end{figure*}

\begin{figure*}[htbp]
\begin{center}
 \mbox{\hspace{-0.2in}
\includegraphics[width=4.5in]{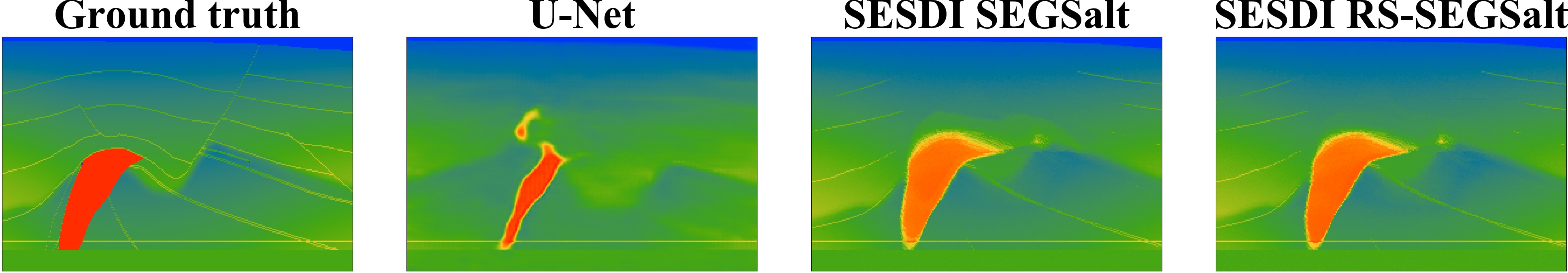}
    }
    \mbox{\hspace{-0.2in}
\includegraphics[width=4.5in]{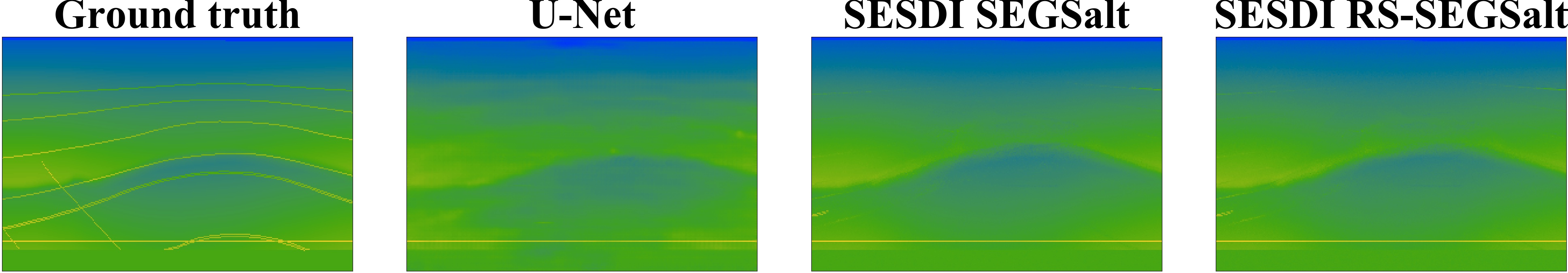}
    }
\mbox{\hspace{-0.2in}
\includegraphics[width=4.5in]{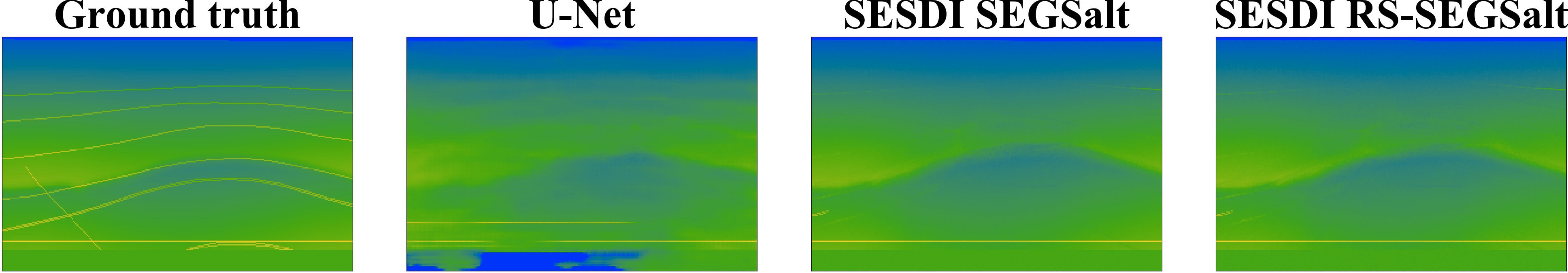}
    }
\mbox{\hspace{-0.2in}
\includegraphics[width=4.5in]{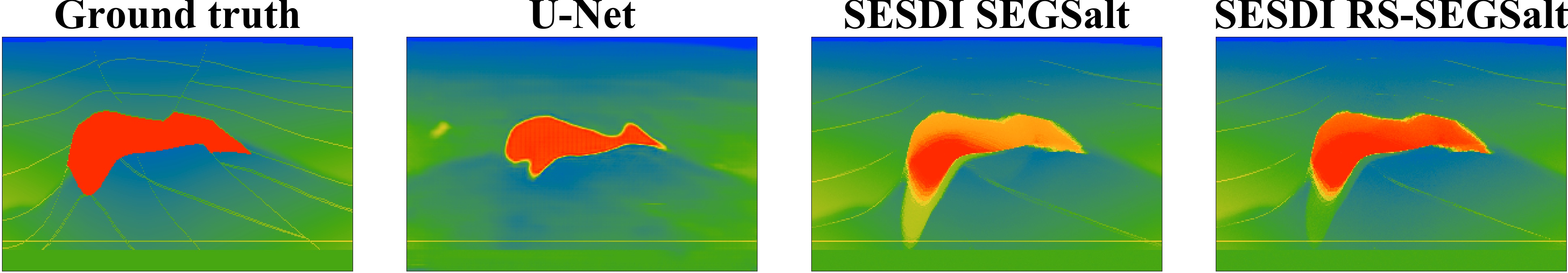}
    }
\mbox{\hspace{-0.2in}
\includegraphics[width=4.5in]{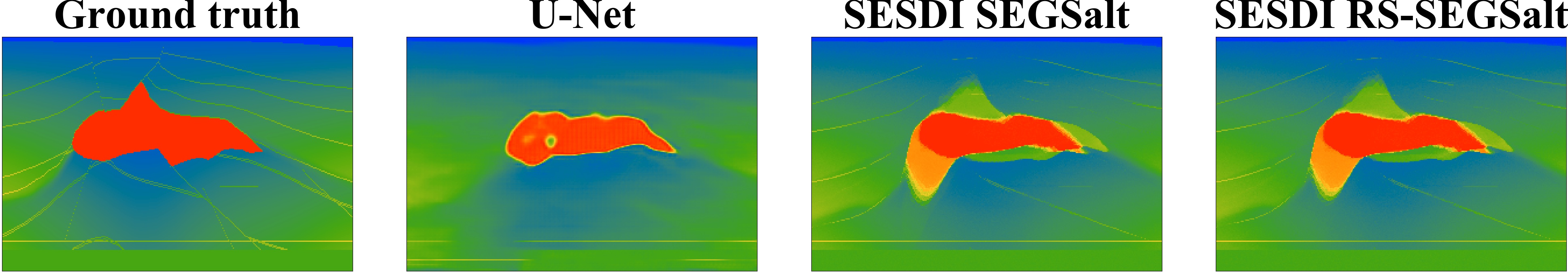}
    }
\mbox{\hspace{-0.2in}
\includegraphics[width=4.5in]{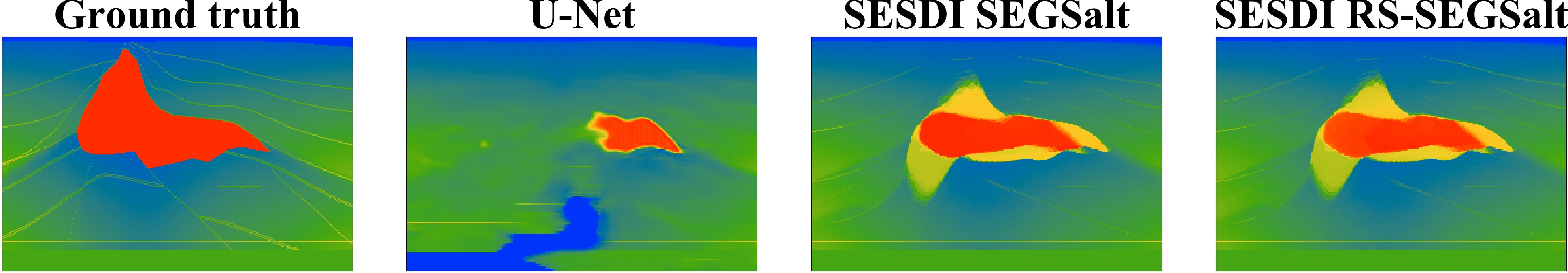}
    }
\mbox{\hspace{-0.2in}
\includegraphics[width=4.5in]{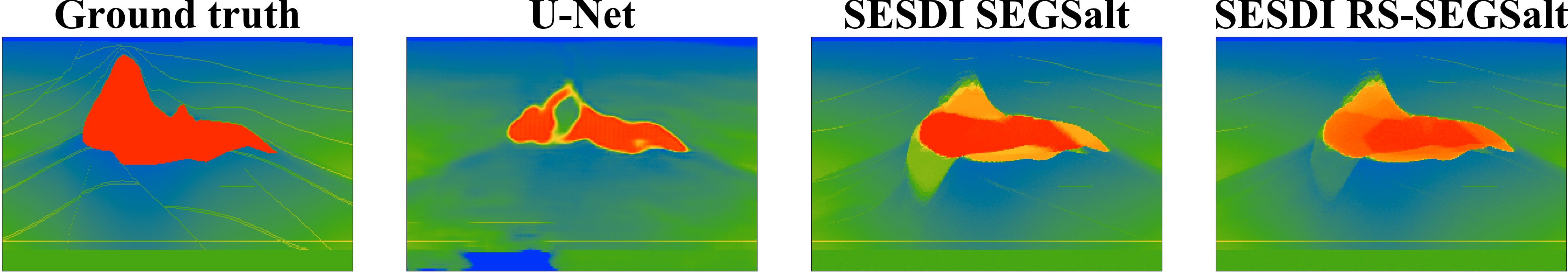}
    }
\mbox{\hspace{-0.2in}
\includegraphics[width=4.5in]{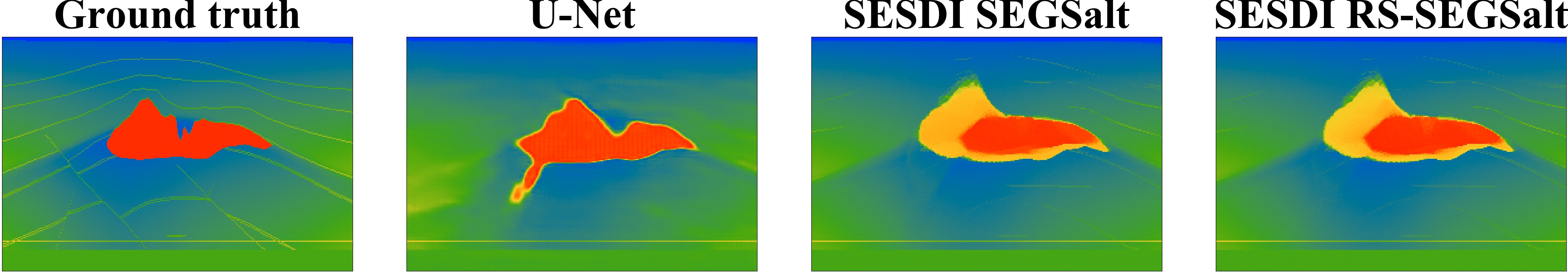}
    }
\mbox{\hspace{-0.2in}
\includegraphics[width=4.5in]{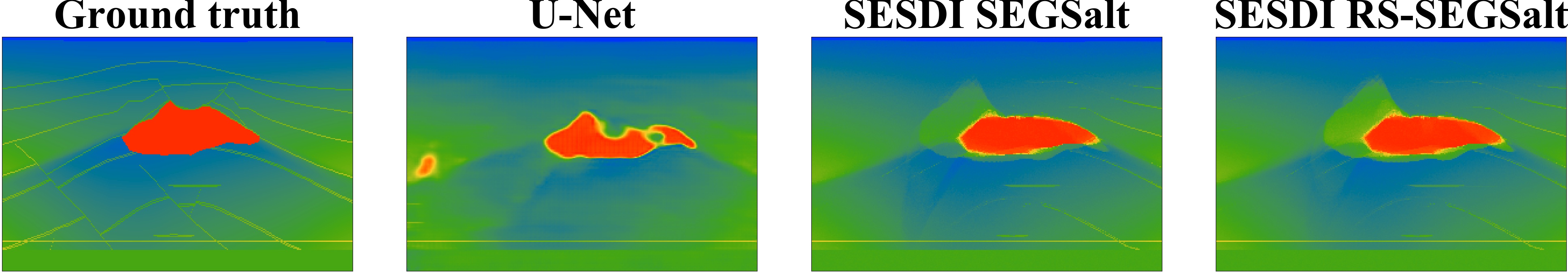}
    }
\mbox{\hspace{-0.2in}
\includegraphics[width=4.5in]{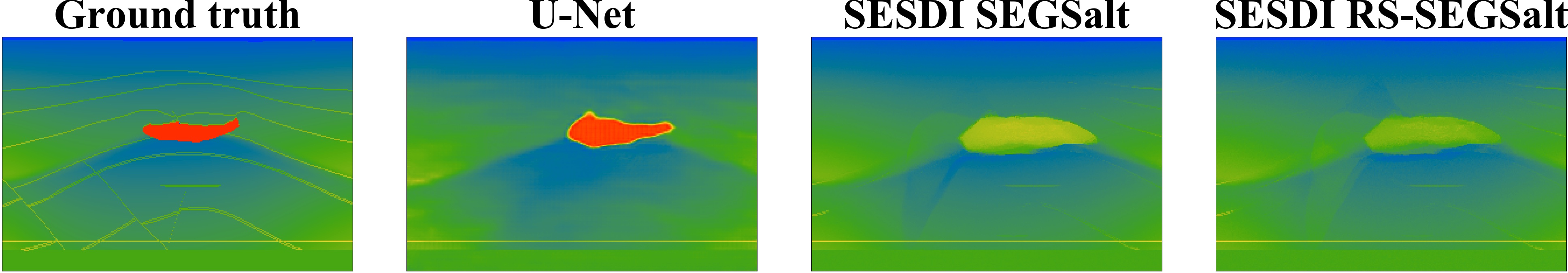}
    }
\mbox{\hspace{-0.25in}
\includegraphics[width=4.5in]{figures/rainbow/bar.pdf}
    }
\end{center}

\vspace{-0.15in}
\caption{Visualization of Prediction on SEGSalt Dataset}\label{fig:salt}
\end{figure*}
\begin{figure*}[htb]
\begin{center}
 \mbox{\hspace{-0.2in}
  \includegraphics[width=.15\textwidth]{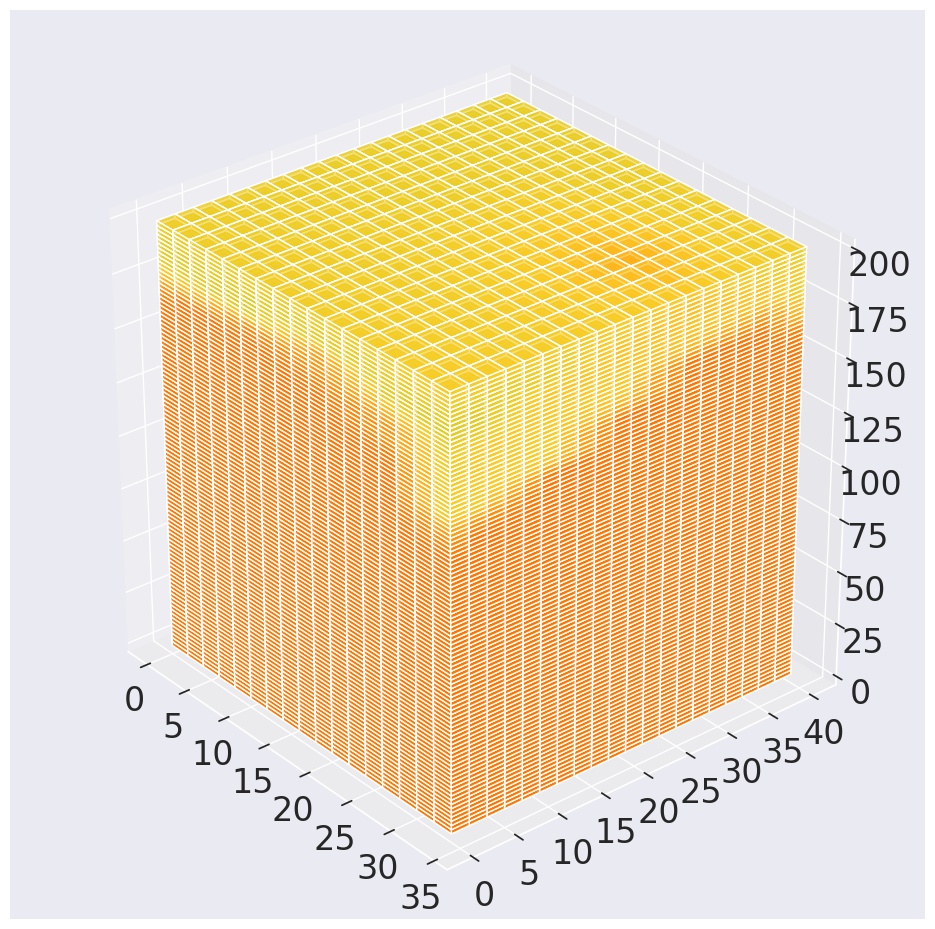}
\includegraphics[width=.15\textwidth]{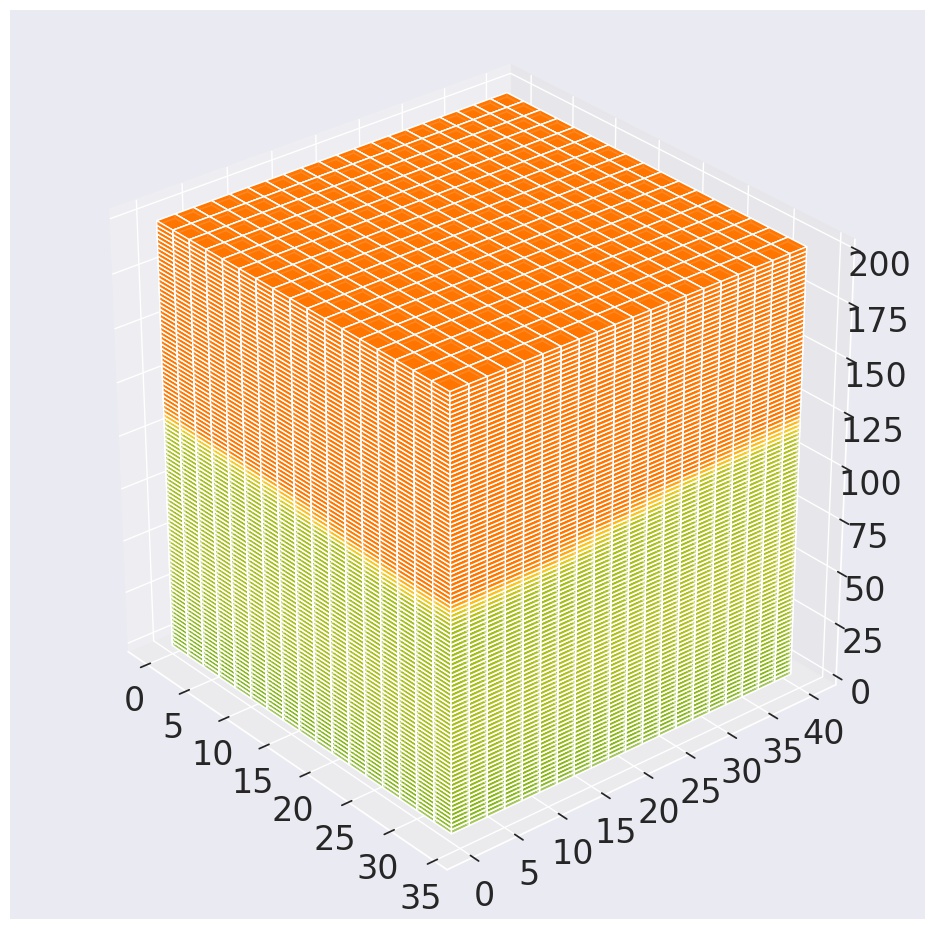}
\includegraphics[width=.15\textwidth]{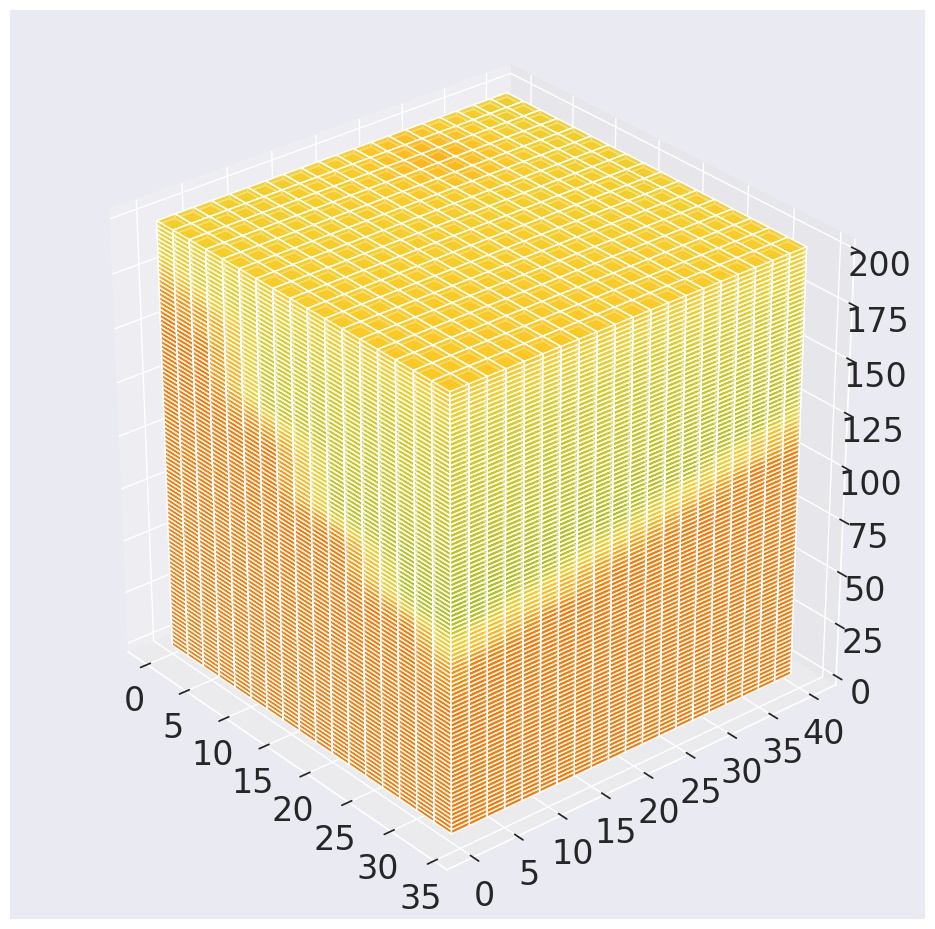}
\includegraphics[width=.15\textwidth]{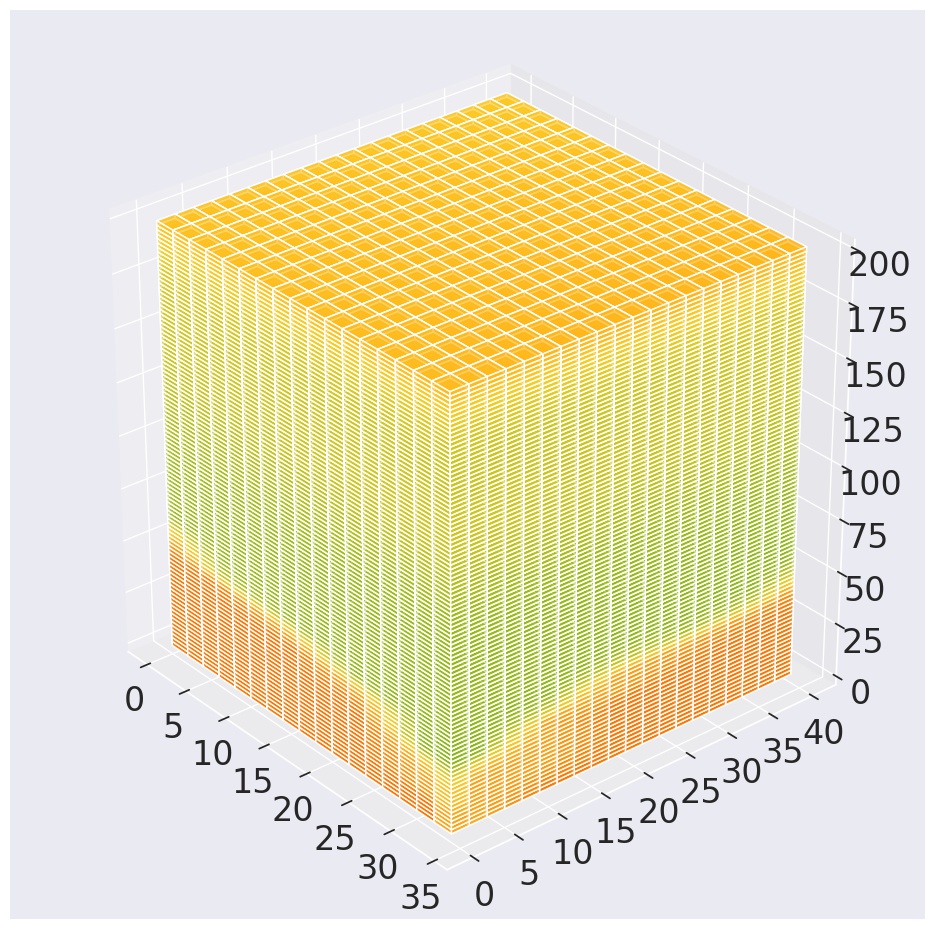}
\includegraphics[width=.15\textwidth]{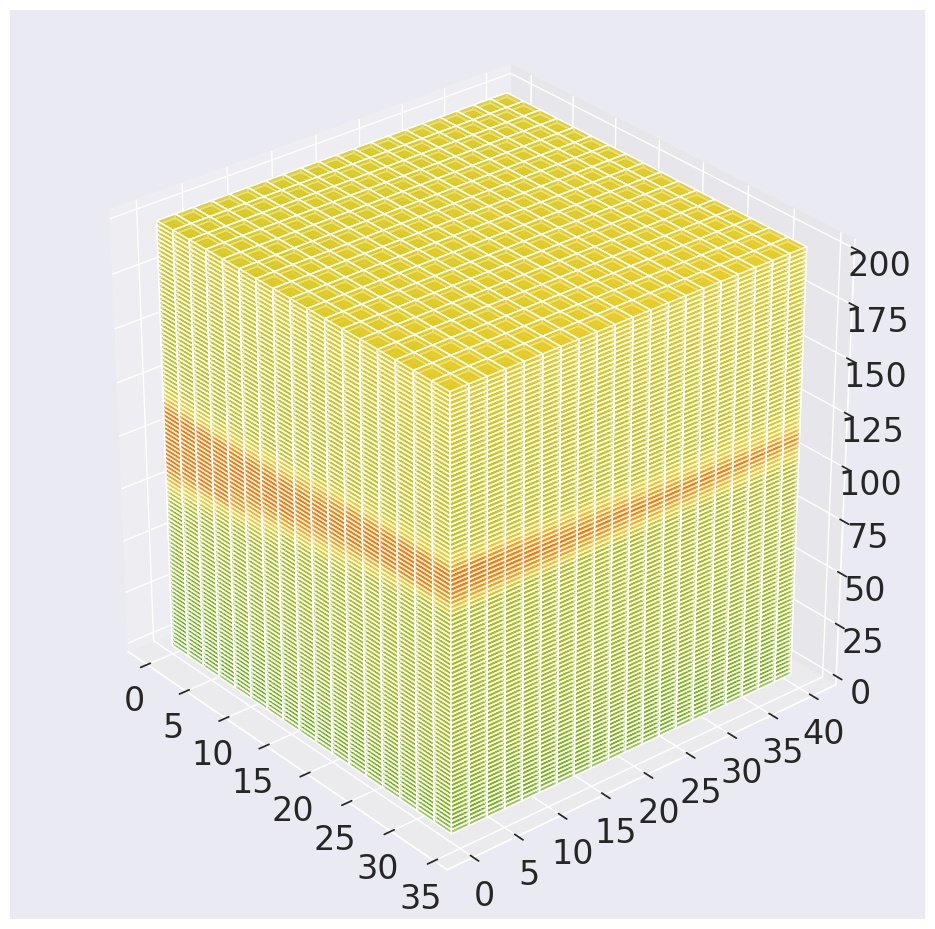}
    }

\mbox{\hspace{-0.2in}
 \includegraphics[width=.15\textwidth]{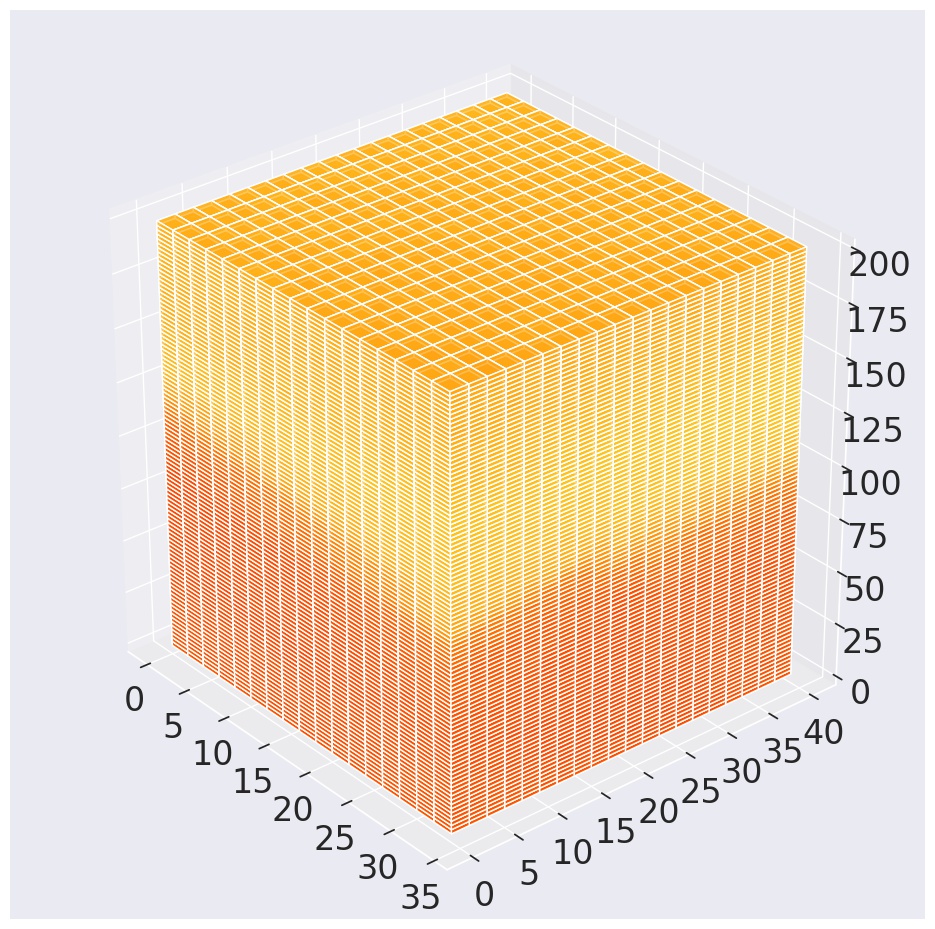}
\includegraphics[width=.15\textwidth]{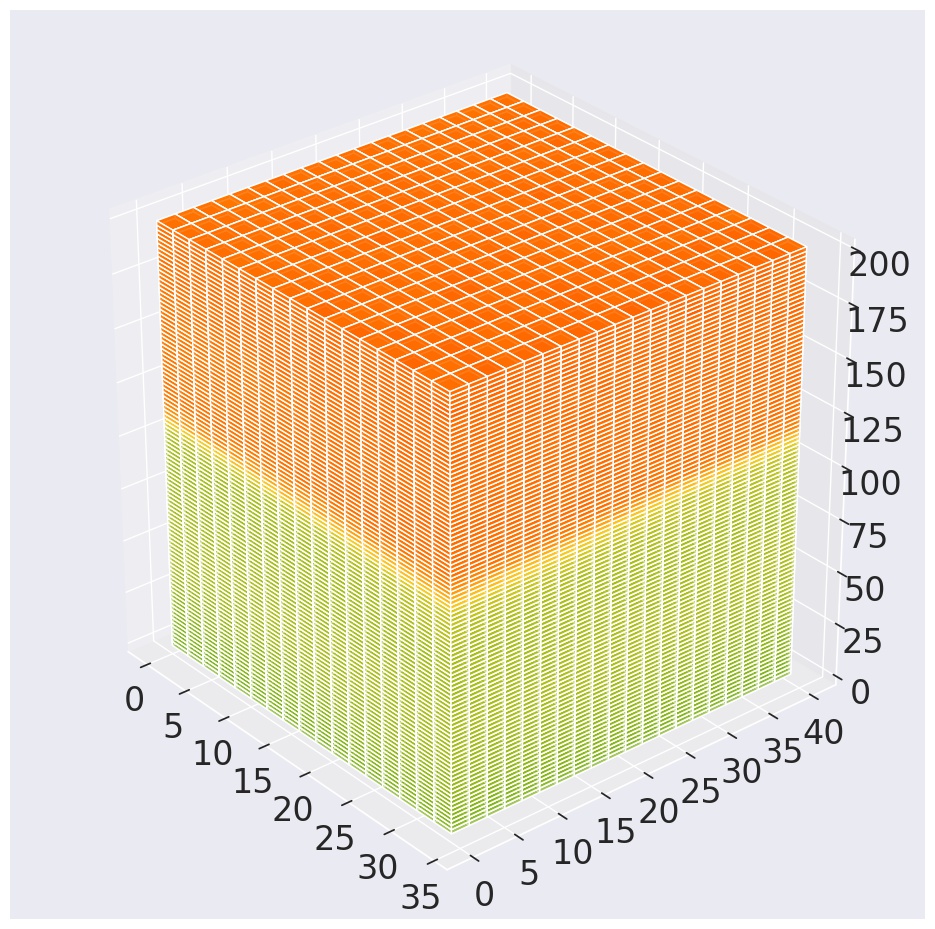}
\includegraphics[width=.15\textwidth]{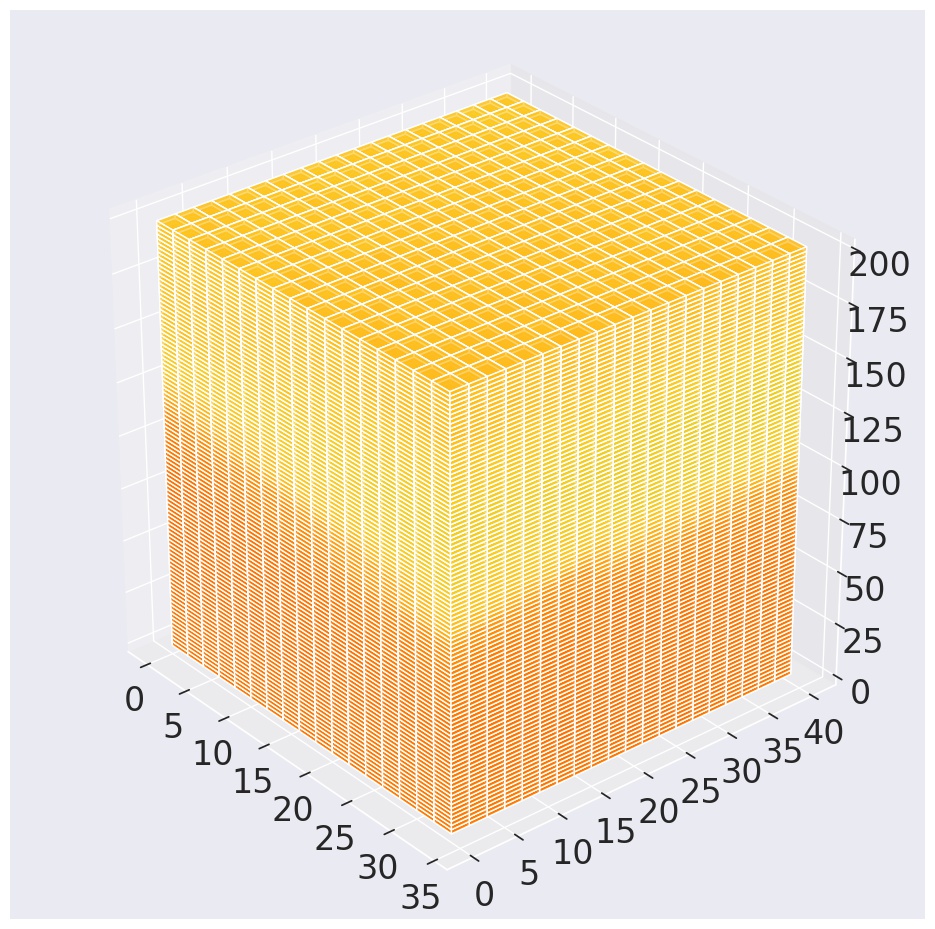}
\includegraphics[width=.15\textwidth]{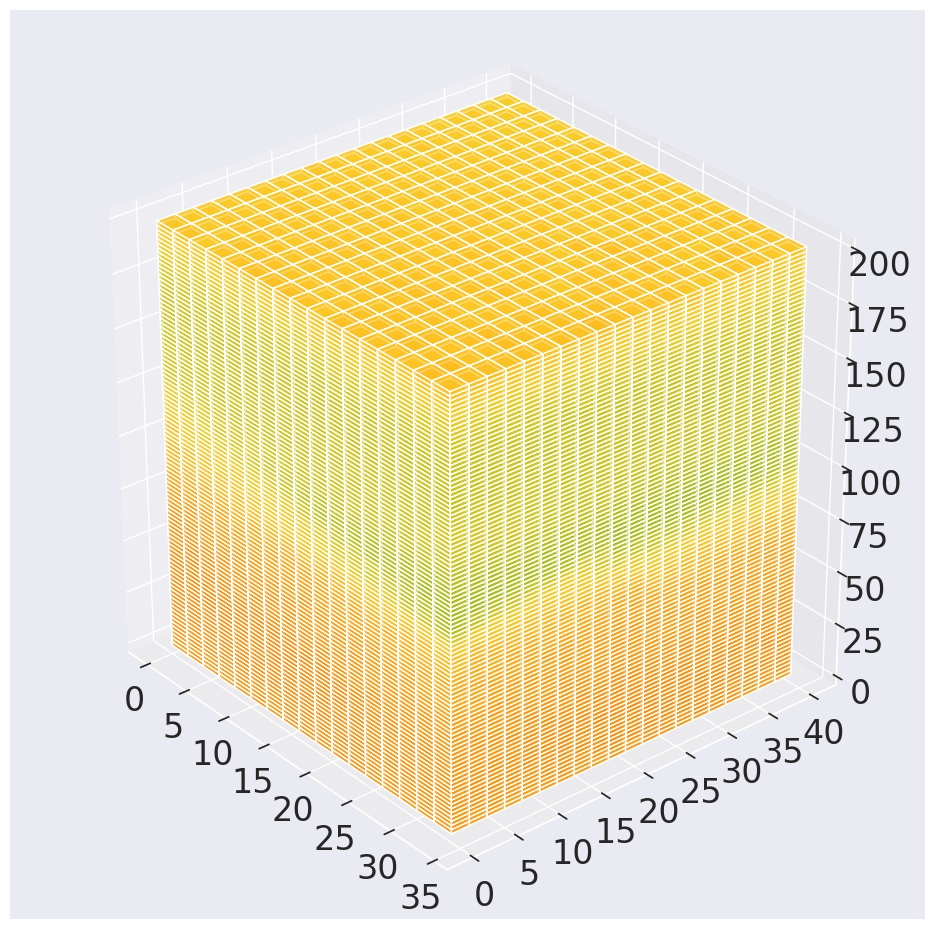}
\includegraphics[width=.15\textwidth]{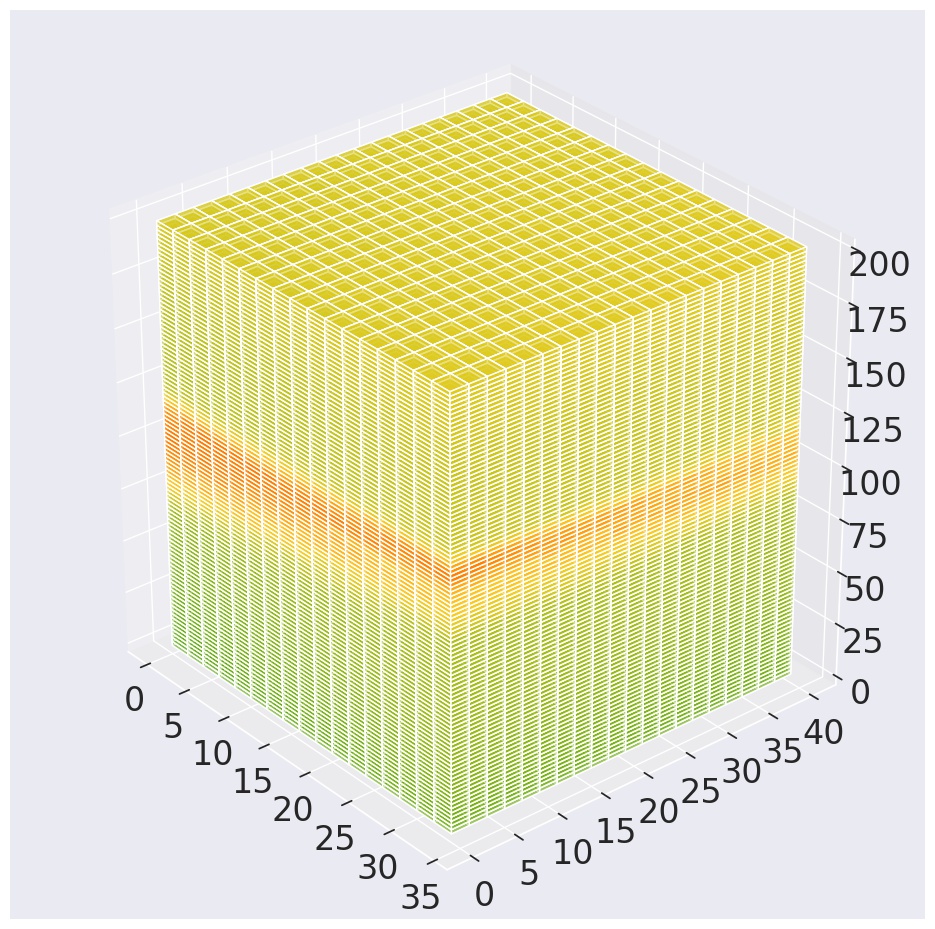}
    }
\mbox{\hspace{-0.2in}
\includegraphics[width=4.5in]{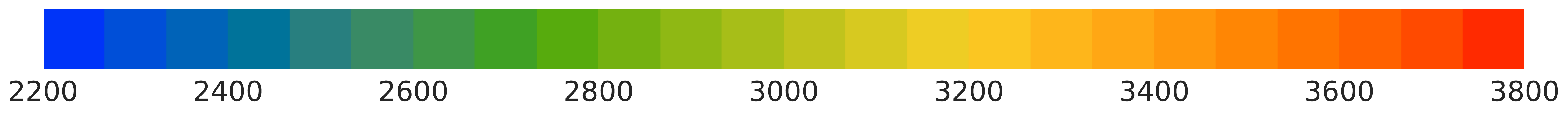}
}

\end{center}

\vspace{-0.15in}
\caption{Visualization of Prediction on GulfM-20 Dataset. Seismic data courtesy of TGS.}\label{fig:nakika100_apx}
\end{figure*}

\subsection{Visualizations}\label{apx:sim_vis}
\begin{figure*}[htbp]
\begin{center}
\includegraphics[width=4.5in]{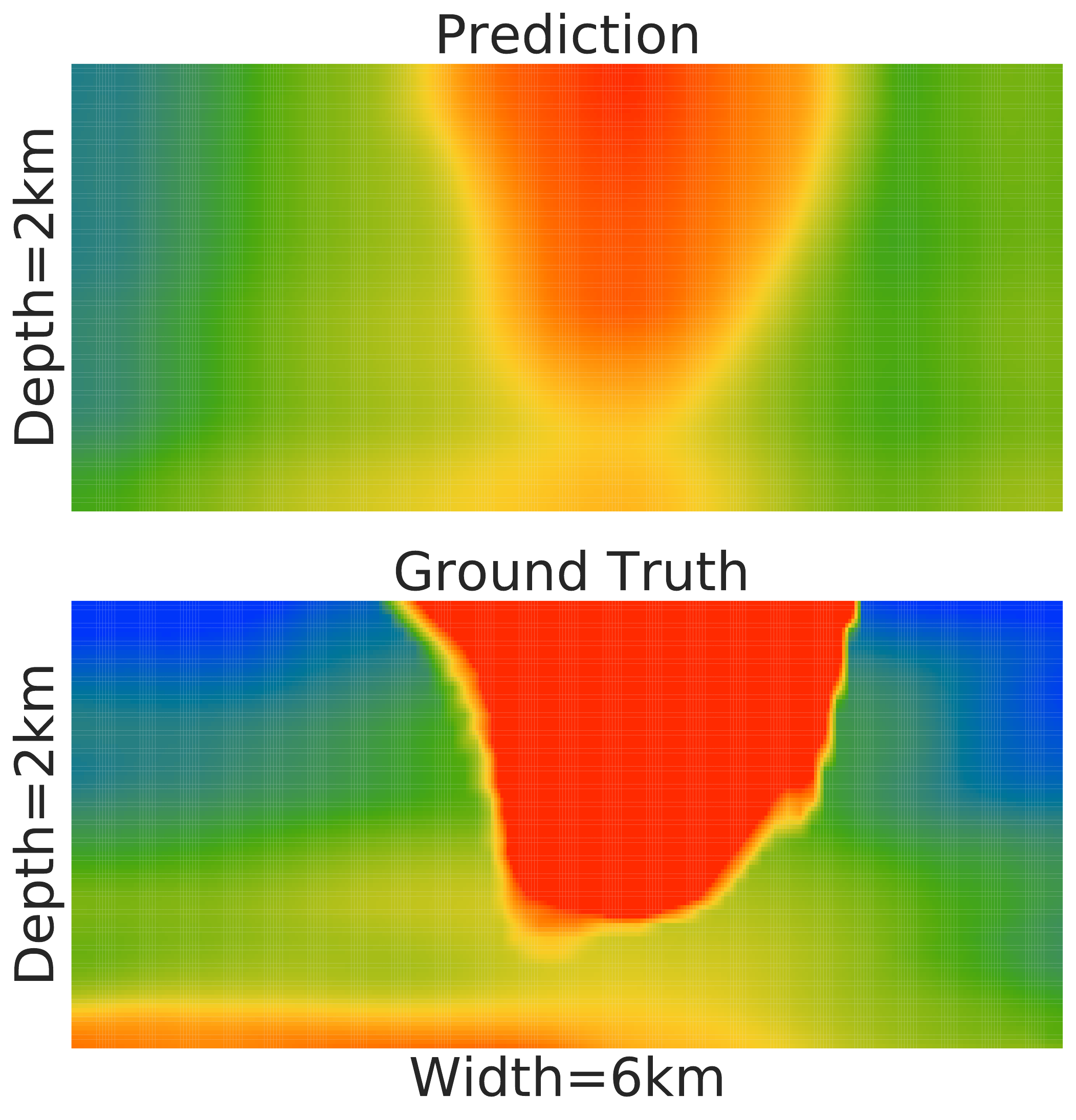}
\end{center}
\vspace{-0.15in}
\caption{Visualization of Prediction on Large Scale VMB in a depth 2km and width 6km area. Seismic data courtesy of TGS.}\label{fig:large_apx}
\end{figure*}

We present the visualization of all predictions in SEGSalt dataset in Figure~\ref{fig:salt}. A sample prediction for GulfM-10 and GulfM-20 dataset is shown in Figure~\ref{fig:nakika50_apx} and Figure~\ref{fig:nakika100_apx}. Here, we choose different color maps for demonstrate the details information in seismic velocities. For the inference on large scale in Figure~\ref{fig:large_apx}, we run the model in the region where both GulfM10 and GulfM20 datasets are randomly sample from.

%\appendix

\end{document}